**Unbox the Black-box: Predict and Interpret YouTube Viewership Using Deep Learning**

Jiaheng Xie
University of Delaware {jxie@udel.edu}

Xiao Liu
University of Utah, {xiao.liu@eccles.utah.edu}

Please send comments to Jiaheng Xie at jxie@udel.edu

**Unbox the Black-box: Predict and Interpret YouTube Viewership Using Deep Learning**

**Abstract:** As video-sharing sites emerge as a critical part of the social media landscape, video viewership prediction becomes essential for content creators and businesses to optimize influence and marketing outreach with minimum budgets. Although deep learning champions viewership prediction, it lacks interpretability, which is required by regulators and is fundamental to the prioritization of the video production process and promoting trust in algorithms. Existing interpretable predictive models face the challenges of imprecise interpretation and negligence of unstructured data. Following the design-science paradigm, we propose a novel Precise Wide-and-Deep Learning (PrecWD) to accurately predict viewership with unstructured video data and well-established features while precisely interpreting feature effects. PrecWD's prediction outperforms benchmarks in two case studies and achieves superior interpretability in two user studies. We contribute to IS knowledge base by enabling precise interpretability in video-based predictive analytics and contribute nascent design theory with generalizable model design principles. Our system is deployable to improve video-based social media presence.

---

## Introduction

Social media is taking up a greater share of consumers' attention and time spent online, among which video-sharing sites, such as YouTube and Vimeo, are quickly overtaking the crown. YouTube alone hosts over 2.6 billion active users and is projected to surpass text- and image-based social media platforms, such as Facebook and Instagram [53]. The soaring popularity of the contents in video format makes video-sharing sites an effective channel to disseminate information and share knowledge.

Content consumption on these social media platforms has been a phenomenon of interest in information systems (IS) and marketing research. Prior work has investigated the impact of digital content on improving sales [18] and boosting awareness of a brand or a product [22]. They also examined the factors that may increase consumption [35] and offered some insights for the design of digital contents [6]. These studies acknowledge that digital content consumption and its popularity are understudied [35]. Our study approaches this domain with an interpretable predictive analytics lens: viewership prediction. Viewership is the metric video-sharing sites use to pay their content creators, defined as the average daily views of a video. While viewership prediction offers immense implications, interpretation elevates such value: evaluating a learned model, prioritizing features, and building trust with domain experts and end users. Therefore, we propose an interpretable machine learning (ML) to predict video viewership (narrative-based long-form videos) and interpret the predictors.

Murdoch et al. [49] show that predictive accuracy, descriptive accuracy, and relevancy form the three pillars of an interpretable ML model. Predictive accuracy measures the model's prediction quality. Descriptive accuracy assesses how well the model describes the data relationships learned by the prediction, or interpretation quality. Relevancy is defined as whether the model provides insight for the domain problem. In this study, both predictive and descriptive accuracy have relevancy to content creators, sponsors, and platforms.

For content creators, the high predictive accuracy of viewership improves the allocation of promotional funds. If the predicted viewership exceeds expectation, the promotional funds can be distributed to less popular videos where content enrichment is insufficient to gain organic views. Meanwhile, high predictive accuracy facilitates trustworthy interpretation which offers effective actions for content creators. Video production requires novel shooting skills and sophisticated communication mindsets for elaborate audio-visual storytelling, which average content creators lack. The interpretation navigates them through how to prioritize the customizable video features.

For sponsors, before their sponsored video is published, high predictive accuracy enables them to estimate the return (viewership) compared to the sponsorship cost. If the return-cost ratio is unsatisfactory, the sponsors could request content enrichment. For platforms, viewership prediction helps control the *influence* of violative videos. Limited by time and resources, the current removal measures are far from being sufficient, resulting in numerous high-profile violative videos infiltrating the public[1]. YouTube projects the percentage of total views from violative videos as a measure of content quality [50]. To minimize the *influence* of violative videos, YouTube could rely on viewership prediction to prioritize the screening of potentially popular videos, among which violative videos can be banned before they reach the public.

Interpretable ML models can be broadly categorized as *post-hoc* and model-based [49]. The general principle of these interpretable methods is to estimate the total effect, defined as the change of the outcome when a feature increases by one unit. *Post-hoc* methods explain a black-box prediction model using a separate explanation model, such as SHAP and LIME [43,52]. However, these standalone explanation models could alter the total effect of the prediction model, since they possess different specifications [43]. Model-based methods address such a limitation by directly interpreting the same prediction model. The cutting-edge model-based interpretable methods include the Generalized Additive Model framework (GAM) and the Wide and Deep Learning framework (W&D). GAM is unable to model higher-order feature interactions and caters to small feature sizes, limiting its applicability in this study. Addressing that, W&D combines a deep learning model with a linear model [13]. However, a few limitations persist for W&Ds. From the prediction perspective, W&Ds are restricted to structured data, constrained by the linear component. In video analytics, only using structured data hampers the predictive accuracy. From the interpretation perspective, W&Ds fall short in producing the precise total effect, defined as the precise change of the prediction when a feature increases one unit. They use the weights of the linear component (main effect) to approximate the total effect of the input on the prediction, even though the main effect and total effect largely differ.

[1] In May 2020, a video called "Plandemic" featured a prominent anti-vaxxer falsely claiming that billionaires were helping to spread the virus to increase use of vaccines. By the time YouTube removed the video, it had already hit 7.1 million views [63]. Other examples are in Appendix 1.

To address the limitations of the existing interpretable methods, we propose a novel model-based interpretable model that learns from both structured features and unstructured video data and produces a precise interpretation, named Precise Wide-and-Deep Learning (PrecWD). This work contributes to data analytics methodology and IS design theory. First, we develop PrecWD that innovatively extends the W&D framework to provide a precise interpretation and perform unstructured data analysis. As our core contribution, the proposed interpretation component can precisely capture the total effect of each feature. A generative adversarial network learns the data distribution and facilitates such an interpretation process. Because of our interpretation component, we are able to add the unstructured component to extend the W&D framework to unstructured data analytics. Empirical evaluations from two case studies indicate that PrecWD outperforms black-box and interpretable models in viewership prediction. We design two user studies to validate the contribution of our precise interpretation component, which indicates PrecWD can provide better interpretability than state-of-the-art interpretable methods. Our feature interpretation results in improved trust and usefulness of the model.

Second, for design science IS research [1,2,21,44,54,60], the successful model design offers indispensable design principles for model development: 1) Our interpretation as well as the user studies suggest generative models can assist the interpretation of predictive models; 2) Our ablation studies indicate raw unstructured data can complement crafted features in prediction. These design principles along with our model provide a "nascent design theory" [27,40,41] that is generalizable to other problem domains. Our new interpretation component can be leveraged by other IS studies to provide precise interpretation for prediction tasks. The user studies can be adopted by IS research to evaluate interpretable ML methods. PrecWD is also a deployable information system for video-sharing sites. It is capable of predicting potentially popular videos and interpreting the factors. Video-sharing sites could leverage this system to actively monitor the predictors and manage viewership and content quality.

## Literature Review

### *Video Viewership Prediction*

This study focuses on YouTube, as it is the most successful video-sharing site since its establishment in 2005 and constitutes the largest share of Internet traffic [42]. We do not use short-form videos (e.g., TikTok and Reels), because most of them use trending songs as the background sound without narratives. Those background songs are the templates provided by the platform that is irrelevant to the video content. The sound of most YouTube videos is the direct description of the video content, for which we designed many narrative-related features, such as sentiment and readability, that are relevant to video production. Viewership is essential for content creators, as it is the key metric used by video-sharing sites to pay them [42]. For the platforms, their user-generated content is constantly bombarded with violative videos, ranging from pornography, copyrighted material, violent extremism, to misinformation. YouTube has recently developed its AI system to prevent violative videos from spreading. This AI system's effectiveness in finding rule-breaking videos is evaluated with a metric called the violative view rate, which is the percentage of views coming from violative videos [50]. This disclosure shows that video viewership is a vital measure for YouTube to track popular videos where credible videos can be approved and violative videos can be banned from publication.

Recognizing the significance of video viewership prediction, prior studies have developed conventional ML and deep learning models [33,39,42,57]. Although reaching sufficient predictive power, these models fail to provide actionable insights for business decision-making due to the lack of interpretability. Studies show that decision-makers exhibit an inherent distrust of automated predictive models, even if they are more accurate than humans [46,47]. Interpretability can increase trust in predictions, expose hidden biases, and reduce vulnerability to adversarial attacks, thus a much-needed milestone for fully harnessing the power of ML in decision-making.

***Interpretability Definition and Value Proposition***

The definition of interpretability varies based on the domain of interest [37]. Two main definitions exist in the area of business analytics, predictive analytics, and social media analytics [16,37,46,47]. One definition is the degree to which a user can trust and understand the cause of a decision [46]. The interpretability of a model is higher if it is easier for a user to trust the model and trace back why a prediction was made. Molnar [47] notes that interpretable ML should make the behavior and predictions of ML understandable and trustworthy to humans. Under the first definition, interpretability is related to how well humans *trust* a model. The second definition suggests "AI is interpretable to the extent that the produced interpretation is able to maximize a user's target performance" [16]. Following this definition, Lee et al. [37] use *usefulness* to measure ML interpretability, as useful models lead to better decision-making performance.

Interpretability brings extensive value to ML and the business world. The most significant one is social acceptance, which is required to integrate algorithms into daily lives. Heider and Simmel [30] show that people attribute beliefs and intentions to abstract objects, so they are more likely to accept ML if their decisions are interpretable. As our society is progressing toward the integration with ML, new regulations have been imposed to require verifiability, accountability, and more importantly, full transparency of algorithm decisions. A key example is the European General Data Protection Regulation (GDPR), which was enforced to provide data subjects the right to an explanation of algorithm decisions [14].

For platforms interested in improving content quality management, an interpretable viewership prediction method not only identifies patterns of popular videos but also facilitates trust and transparency in their ML systems. For content creators, it takes significant time and effort to create such content. An interpretable viewership prediction method can recommend optimal prioritization of video features in the limited time.

### *Interpretable Machine Learning Methods*

We develop a taxonomy of the extant interpretable methods in Table 1 based on the data types, the type of algorithms (i.e., model-based and *post-hoc*), the scope of interpretation (i.e., instance level and model level), and how they address interpretability. [Insert Table 1 here]

Various forms of data have been used to train and develop interpretable ML methods, including tabular [28,61], image [26,56], and text [5]. The scope of interpretation can be at either instance- or model-level. An interpretable ML method can be either embedded in the neural network (model-based) or applied as an external model for explanation (*post-hoc*) [49]. *Post-hoc* methods build on the predictions of a black-box model and add *ad-hoc* explanations [26,56]. Any interpretable ML algorithm that directly interprets the original prediction model falls into the model-based category [5,10,28]. For most model-based algorithms, any change in the architecture needs alteration in the method or hyperparameters of the interpretable algorithm.

*Post-hoc* methods can be backpropagation- or perturbation-based. The backpropagation-based methods rely on gradients that are backpropagated from the output layer to the input layer [75]. Yet, the most widely used are the perturbation-based methods. These methods generate explanations by iteratively probing a trained ML model with different inputs. These perturbations can be on the feature level by replacing certain features with zeros or random counterfactual instances. Shapley Additive Explanations (SHAP) is the most popular perturbation-based *post-hoc* method. It probes feature correlations by removing features in a game-theoretic framework [43].

The Local Interpretable Model-agnostic Explanations (LIME) is another common perturbation-based method [52]. For an instance and its prediction, simulated randomly sampled data around the neighborhood of the input instance are generated. An explanation model is trained on this newly created dataset of perturbed instances to explain the prediction of the black-box model. SHAP and LIME are both feature additive methods that use an explanation model that is a linear function of binary variables: $g(z') = \phi_0 + \sum_{i=1}^{M} \phi_i z_i'$ [43]. The prediction of this explanation model matches the prediction model $f(x)$ [43]. Eventually $\phi_i$ can explain the attribution of each feature to the prediction.

These *post-hoc* methods have pitfalls. Laugel et al. [36] showed that *post-hoc* models risk having explanations resulted from the artifacts learned by the explanation model instead of the actual knowledge from the data. This is because the most popular *post-hoc* models SHAP and LIME use a separate explanation model $g(z')$ to explain the original prediction model $f(x)$ [43]. The model specifications of $g(z')$ are fundamentally different from $f(x)$. The feature contributions $\phi_i$ from $g(z')$ are not the actual feature effect of the prediction model $f(x)$. Therefore, the magnitude and direction of the total effect could be misinterpreted. Slack et al. [59] revealed that SHAP and LIME are vulnerable, because they can be arbitrarily controlled. Zafar and Khan [74] reported that the random perturbation that LIME utilizes results in unstable interpretations, even in a given model specification and prediction task.

Addressing the limitations of *post-hoc* methods, model-based interpretable methods have a self-contained interpretation component that is faithful to the prediction. The model-based methods are usually based on two frameworks: the Generalized Additive Model framework (GAM) and the Wide and Deep Learning framework (W&D). GAM's outcome variable depends linearly on the smooth functions of predictors, and the interest focuses on inference about these smooth functions. However, for prediction tasks with many features, GAMs often require millions of decision trees to provide accurate results using additive algorithms. Also, depending on the model architecture, over-regularization reduces the accuracy of GAM. Many methods have improved GAMs: GA2M was proposed to improve the accuracy while maintaining the interpretability of GAMs [10]; NAM learns a linear combination of neural networks where each of them attends to a single feature, and each feature is parametrized by a neural network [5]. These networks are trained jointly and can learn complex-shape functions. Interpreting GAMs is easy as the impact of a feature on the prediction does not rely on other features and can be understood by visualizing its corresponding shape function. However, GAMs are constrained by the feature size, because each feature is assumed independent and trained by a standalone model. When the feature size is large and feature interactions exist, GAMs struggle to perform well [5].

The W&D framework addresses the low- and high-order feature interactions in interpreting the importance of features [13]. W&Ds are originally proposed to improve the performance of recommender systems by combining a generalized linear model (wide component) and a deep neural network (deep component) [13]. Since the generalized linear model is interpretable, as noted in [13], soon this model has been recognized or used as an interpretable model in many applications [12,28,62,68]. The wide component produces a weight for each feature, defined as the main effect, that can interpret the prediction. The deep component models high-order relations in the neural network to improve predictive accuracy. Since W&D, two categories of variants have emerged. The first category improves the predictive accuracy of W&D. Burel et al. [9] leveraged CNN in the deep component to identify information categories in crisis-related posts. Han et al. [29] used a CRF layer to merge the wide and deep components and predict named entities of words. The second category improves the interpretability of W&D. Chai et al. [12] leveraged W&D for text mining interpretation. Guo et al. [28] proposed piecewise W&D where multiple regularizations are introduced to the total loss function to reduce the influence of the deep component on the wide component so that the weights of the wide component are closer to the total effect. Tsang et al. [62] designed an interaction component in W&D. This method was used to interpret the statistical interactions of housing price and rental bike count predictions.

The W&D and its variants still fall short in two aspects. First, W&D uses the learned weights of the wide component $\boldsymbol{w}^{\mathrm{T}}$ (main effect) to interpret the prediction. $\boldsymbol{w}^{\mathrm{T}}$only reflects the linear component which is only a portion of the entire prediction model. Consequently, $\boldsymbol{w}^{\mathrm{T}}$ is not the total effect of the joint model. For instance, the weight $w_1$ for feature $x_1$ does not imply that if $x_1$ increases by one unit, the viewership prediction would increase $w_1$. The real feature interpretation for $x_1$ cannot be precisely reflected in $w_1$. This imprecise interpretation also occurs in *post-hoc* methods and GAMs, due to their interpretation mechanisms. *Post-hoc* methods, such as SHAP and LIME, use an independent explanation model as a proxy to explain the original prediction model. This separate explanation mechanism inherently cannot directly nor precisely interpret the original prediction model. GAMs interpret each feature independently using a standalone model, thus cannot interpret the precise feature effect when all the features are used together. Precise total effect is critical, as it affects the weight (importance) of feature effects, which determines the feature importance order and effect direction. Correct feature importance order is essential for content creators to know which feature to prioritize. Given limited time, such order informs content creators of the prioritization of the work process. In addition, $\boldsymbol{w}^{\mathrm{T}}$ is constant for all values of $\boldsymbol{x}$, which assumes the feature effect is insensitive to changes of feature value. This assumption does not hold in real settings. For instance, when a video is only minutes long, increasing one minute would significantly impact its predicted viewership. When a video is hours long, increasing one minute does not have a visible effect on predicted viewership.

Second, unstructured data are not compatible with the W&D framework. The existing W&D framework enforces the wide component and the deep component to share inputs and be trained jointly, so that the wide component can interpret the deep component. The wide component in W&D is a linear model: $f^{W}(\boldsymbol{x}) = \boldsymbol{w}^{\mathrm{T}}\boldsymbol{x} + b$, where $\boldsymbol{x} = [x_1, x_2, \dots, x_d]$ is a vector of $d$ features, including raw input features and product-transformed features [13]. The raw input features are numeric, including continuous features and vectors of categorical features [13]. The product transformation is $\phi_k(\boldsymbol{x}) = \prod_i x_i^{\xi_{ki}}$ $(\xi_{ki} \in \{0,1\})$, where $\boldsymbol{x}$ is the raw numeric input feature $[x_1, x_2, \dots, x_d]$, and $\xi_{ki}$ indicates whether the $i$-th feature appears in the $k$-th transformation. Both raw input features and product-transformed features are structured data. This is due to the structured nature of the linear model $\boldsymbol{w}^{\mathrm{T}}\boldsymbol{x} + b$. It is incapable of processing unstructured videos in this study, thus significantly limiting W&D's performance in unstructured data analytics.

***Generative Models for Synthetic Sampling***

In order to calculate the precise total effect, we develop a novel model-based interpretation method, which we will detail in the Proposed Approach section. Learning the data distribution is critical for the precise total effect with a model-based interpretation method, which can be facilitated by synthetic sampling with generative models.

Early forms of generative models date back to Bayesian Networks and Helmholtz machines. Such models are trained via an EM algorithm using variational inference or data augmentation [19]. Bayesian Networks require the knowledge of the dependency between each feature pair, which is useful for cases with limited features that have domain knowledge. When the feature size is large, constructing feature dependencies is infeasible and leads to poor performance. Recent years have seen developments in deep generative models. The emerging approaches, including Variational Autoencoders (VAEs), diffusion models, and Generative Adversarial Networks (GANs), have led to impressive results in various applications. Unlike Bayesian Networks, deep generative models do not require the knowledge of feature dependencies.

VAEs use an encoder to compress random samples into a low-dimensional latent space and a decoder to reproduce the original sample [55]. VAEs use variational inferences to generate an approximation to a posterior distribution. Diffusion models include a forward diffusion stage and a reverse diffusion stage. In the forward diffusion stage, the input data is gradually perturbed over several steps by adding Gaussian noise. In the reverse stage, a model is tasked at recovering the original input data by learning to gradually reverse the diffusion process, step by step [15]. GANs are a powerful class of deep generative models consisting of two networks: a generator and a discriminator. These two networks form a contest where the generator produces high-quality synthetic data to fool the discriminator, and the discriminator distinguishes the generator's output from the real data. Deep learning literature suggests that the generator could learn the precise real data distribution as long as those two networks are sufficiently powerful [19,73]. The resulting model is a generator that can closely approximate the real distribution.

Although VAEs, diffusion models, and GANs are emergent generative models, GANs are preferred in this study. VAEs optimize the lower variational bound, whereas GANs have no such assumption. In fact, GANs do not deal with any explicit probability density estimation. The requirement of VAEs to learn explicit density estimation hinders their ability to learn the true posterior distribution. Consequently, GANs yield better results than VAEs [7]. Diffusion models are mostly tested in computer vision tasks. Its effectiveness in other contexts lacks strong evidence. Besides, diffusion models suffer from long sampling steps and slow sampling speed, which limits their practicality [15]. Without careful refinement, diffusion models may also introduce inductive bias, which is undesired compared to GANs.

## The Proposed Approach

### *Problem Formulation*

Let $\boldsymbol{V}$ denote a set of unstructured raw videos $(v_1, \ldots, v_N)$. Let $\boldsymbol{X}$ denote the structured video features $(X_1, \ldots, X_M)$. The feature values of a video $v_i$ are represented by $\boldsymbol{x}_i = (x_{i,1}, \ldots, x_{i,M})$. The input to our model is $\boldsymbol{X}_{all} = (\boldsymbol{X}, \boldsymbol{V})$. Viewership is operationalized as the average daily views of a video ($ADV$), computed as the view counts to date divided by the video age in days. A longitudinal study about YouTube viewership shows that the average daily views are stable in the long term [71]. In other words, "older videos are not forgotten as they get older" [71]. Many other studies have also used daily views as a variable of interest [23,34,71]. The intended practical use of our method is also to predict long-term viewership. For this purpose, ADV is an appropriate measure. The $ADV$ of video $v_i$ is denoted as $adv_i$. Our objective is to learn a model $F$ to predict $adv_i$ where $adv_i = F(\boldsymbol{x}_i, v_i)$, and interpret the precise total effect of each feature $X_j$ on the output $adv$ ($\Delta ADV / X_j$) in the given model and feature setting.

***The PrecWD Model for Video Viewership Prediction and Interpretation***

The proposed model builds upon the state-of-the-art W&D framework while addressing two challenges: 1) W&D cannot offer the precise total effect and its dynamic changes; 2) W&D can only process structured data. We propose PrecWD, consisting of the following subcomponents.

*Piecewise Linear Component*

Each feature $X_j$ captures a different aspect of a video. Within each feature, heterogeneity between different values exists. It is essential to consider the homogeneity among similar feature values and the heterogeneity across different feature values. Specifically, we need to differentiate the varied feature effects when the feature is at different values. We leverage a piecewise linear function in the linear component, which is adopted from [28]. For the $j$-th feature, let $\beta_j = \max\{x_{i,j} | i = 1, \ldots, N\}$ and $\delta_j = \min\{x_{i,j} | i = 1, \ldots, N\}$. We partition each feature into $\gamma_j$ intervals: $[\varphi_j^0, \varphi_j^1]$, …, $[\varphi_j^{\gamma_j - 1}, \varphi_j^{\gamma_j}]$, where $\varphi_j^k = \delta_j + \frac{k}{\gamma_j}(\beta_j - \delta_j)$. The piecewise feature vector for the $i$-th data point is:

$$\boldsymbol{\Phi}_i = \left(\phi_{i,1}^1, \ldots, \phi_{i,1}^{\gamma_1}, \ldots, \phi_{i,j}^1, \ldots, \phi_{i,j}^{\gamma_j}, \ldots, \phi_{i,M}^1, \ldots \phi_{i,M}^{\gamma_M}\right)^{\mathrm{T}} \quad (1)$$

where $\phi_{i,j}^{k} = \begin{cases} 1, x_{i,j} > \varphi_j^k \ . \\ \frac{x_{i,j}-\varphi_j^{k-1}}{\varphi_j^k-\varphi_j^{k-1}}, \varphi_j^{k-1} \le x_{i,j} \le \varphi_j^k \\ 0, \text{otherwise}. \end{cases}$. This piecewise vector $\boldsymbol{\Phi}_i$ is then fed into a linear model:

$$y_i^{\mathcal{L}} = (\boldsymbol{w}^{\mathcal{L}})^{\mathrm{T}} \boldsymbol{\Phi}_i + b^{\mathcal{L}} \tag{2}$$

where $(\boldsymbol{w}^{\mathcal{L}})^{\mathrm{T}}$ is the weight in the linear component, $b^{\mathcal{L}}$ is the bias, and $y_i^{\mathcal{L}}$ is the output.

*Attention-based Second-Order Component*

Prior studies suggest explicitly modeling feature interactions improves predictive accuracy [58]. In parallel with the piecewise linear component, we include an attention-based second-order component to model the feature interactions. The input to this component is $\boldsymbol{X}$. For each video feature $\boldsymbol{x}_i$, the interaction term of $x_{i,j}$ and $x_{i,j'}$ is denoted as $s_{i,(j,j')} = x_{i,j} \cdot x_{i,j'}$. Each interaction term has a parameter. A set of $M$ features will generate $M^2$ interaction terms. This will cause the learnable parameters to grow quadratically as the feature size increases. To prevent such a quadratic growth and optimize computational complexity, we add an attention mechanism in the second-order component where the number of interactions is fixed. The attention-based component could scale to a large number of interactions while salient interaction terms still stand out. The attention mechanism assigns a score $a_{i,(j,j')}$ to each interaction term $s_{i,(j,j')}$.

$$a_{i,(j,j')} = \frac{\exp\left(u_{i,(j,j')} \cdot h^{\mathcal{A}}\right)}{\sum_{j=1,j'=1}^{M,M} \exp\left(u_{i,(j,j')} \cdot h^{\mathcal{A}}\right)} \tag{3}$$

$$u_{i,(j,j')} = \tanh\left(w^{\mathcal{A}} \cdot s_{i,(j,j')} + b^{\mathcal{A}}\right) \tag{4}$$

where $w^{\mathcal{A}}$ and $b^{\mathcal{A}}$ are learnable parameters that are shared for all $(j, j')$. The attention score $a_{i,(j,j')}$ is used to weigh the interaction terms. The output is as follows, where $w^{\mathcal{S}}$ is a learnable weight and $\sum_j^M \sum_{j'}^M a_{i,(j,j')} s_{i,(j,j')}$ is the weighted sum of salient interaction terms.

$$y_i^{\mathcal{S}} = w^{\mathcal{S}} \sum_{j}^{M} \sum_{j'}^{M} a_{i,(j,j')} s_{i,(j,j')} \tag{5}$$

*Nonlinear Higher-Order Component*

The third component is a deep neural network that captures higher-order effects. The number of hidden layers is determined using a grid search in the empirical analyses. The purpose of the higher-order component is to leverage deep learning to improve predictive accuracy. Without loss of generality, for the $i$-th video, each hidden layer computes:

$$\boldsymbol{a}_{i,(l+1)} = f\left(\boldsymbol{W}_{(l)}^{\mathcal{H}} \boldsymbol{a}_{i,(l)} + \boldsymbol{b}_{(l)}^{\mathcal{H}}\right) \tag{6}$$

where $l$ is the layer number. $f$ is the ReLU. $\boldsymbol{a}_{i,(l)}$, $\boldsymbol{b}_{(l)}^{\mathcal{H}}$, and $\boldsymbol{W}_{(l)}^{\mathcal{H}}$ are the input, bias, and weight at the $l$-th layer. Therefore, the input of the first layer is the feature vector (i.e., $\boldsymbol{a}_{i,(1)} = \boldsymbol{x}_i$). The output of this component is given by

$$y_i^{\mathcal{H}} = (\boldsymbol{w}^{\mathcal{H}})^{\mathrm{T}} \boldsymbol{a}_{i,(L+1)} \tag{7}$$

where $(\boldsymbol{w}^{\mathcal{H}})^{\mathrm{T}}$is the learnable weight and $L$ is the number of layers.

*Unstructured Component*

The W&D enforces the wide component and deep component to share inputs so that the wide component can interpret the deep component. Since the wide component can only analyze structured data, the W&D is also restricted to structured data. However, videos are unstructured by nature. Obtaining high predictive accuracy in video analytics demands the capacity to process unstructured data. We relax the restraint of the subcomponents sharing inputs, because our proposed interpretation component could offer precise total effects without soliciting dependencies on the subcomponents. We extend the W&D with an unstructured component. The input to our model is $\boldsymbol{X}_{all} = (\boldsymbol{X}, \boldsymbol{V})$. The structured and practically meaningful features are grouped in $\boldsymbol{X}$, which is fed to the previous three components. Unstructured video data are grouped in $\boldsymbol{V}$ that is fed into the unstructured component. Two approaches can incorporate the unstructured data into our model: either use a representation learning model to learn hidden features for $\boldsymbol{V}$, then mix those hidden features with $\boldsymbol{X}$; or design an unstructured component that can directly process raw videos and separate the unstructured effect from the structured effect. Both approaches are viable with our model, but we opt to the latter approach because the learned hidden features of $\boldsymbol{V}$ are not human-understandable. Therefore, $\boldsymbol{V}$ is more helpful in improving prediction accuracy rather than interpretability. Separating it from $\boldsymbol{X}$ could ensure that $\boldsymbol{X}$ can be cleaned up with the carefully designed, understandable, and practically meaningful video features. When using the wide component to process $\boldsymbol{X}$, it will be very clear to see the main effect which can be compared with our total effect. We devise a hybrid VGG-LSTM architecture for processing videos, shown in Equations 8-11. A VGG-16 architecture is designed to process the video frames, and an LSTM layer is added on the top for frame-by-frame sequence processing. The last LSTM cell summarizes the video information for the $i$-th video $v_i$.

$$\tilde{\boldsymbol{y}}_i^{\mathcal{U}} = \mathrm{VGG}(v_i) \tag{8}$$

$$\begin{pmatrix} \boldsymbol{i}^{\mathcal{U}} \\ \boldsymbol{f}^{\mathcal{U}} \\ \boldsymbol{o}^{\mathcal{U}} \\ \boldsymbol{g}^{\mathcal{U}} \end{pmatrix} = \begin{pmatrix} \sigma \\ \sigma \\ \sigma \\ \tanh \end{pmatrix} \boldsymbol{W}^{\mathcal{U}} \begin{pmatrix} \boldsymbol{h}_{t-1}^{\mathcal{U}} \\ \tilde{\boldsymbol{y}}_{i,t}^{\mathcal{U}} \end{pmatrix} \tag{9}$$

$$\boldsymbol{h}_t^{\mathcal{U}} = \boldsymbol{o}^{\mathcal{U}} \odot \tanh\left(\boldsymbol{f}^{\mathcal{U}} \odot \boldsymbol{c}_{t-1}^{\mathcal{U}} + \boldsymbol{i}^{\mathcal{U}} \odot \boldsymbol{g}^{\mathcal{U}}\right) \tag{10}$$

$$y_i^{\mathcal{U}} = \mathrm{Softmax}\left(\mathrm{MLP}\left(\boldsymbol{h}_T^{\mathcal{U}}\right)\right) \tag{11}$$

where $(\boldsymbol{i}^{\mathcal{U}}, \boldsymbol{f}^{\mathcal{U}}, \boldsymbol{o}^{\mathcal{U}}, \boldsymbol{g}^{\mathcal{U}})$ are the gates, $\boldsymbol{h}_t^{\mathcal{U}}$ is the hidden state, and $y_i^{\mathcal{U}}$ is the output.

*Precise Interpretation Component*

Our core contribution lies in the precise interpretation component. Our primary focus is to offer the precise total effect of viewership prediction. PrecWD predicts the outcome variable using

$$\widehat{ADV} = \mathrm{ReLU}(\beta + \alpha_1 X_1 + \ldots + \alpha_M X_M + \mathcal{S}(X_1, \ldots, X_M) + \mathcal{H}(X_1, \ldots, X_M) + \mathcal{U}(v)) \tag{12}$$

where $\beta + \alpha_1 X_1 + \ldots + \alpha_M X_M$ denotes the main effect, $\mathcal{S}(X_1, \ldots, X_M)$ denotes the second-order effect, $\mathcal{H}(x_1, \ldots, x_n)$ denotes the higher-order effect, and $\mathcal{U}(v)$ denotes the unstructured effect. We use ReLU because viewership is non-negative. Take feature $X_1$ as the illustration example, the existing W&Ds would use $\alpha_1$ to approximate the effect of $X_1$ on the prediction, which is different from the actual total effect of $X_1$ that equals to the change of $\widehat{ADV}$ when $X_1$ increases by one unit.

In order to model the precise total effect and its dynamic changes, we predict the total effect of each feature at every value. The precise total effect of $X_1$ when $X_1 = c$ is

$$\begin{aligned}
&\Delta\widehat{ADV}(X_1 = c|X_2, \dots, X_M) \\
&\quad = \mathrm{ReLU}\big(\beta + \alpha_1(c+1) + \cdots + \alpha_M X_M + \mathcal{S}(X_1 = c+1, \dots, X_M) + \mathcal{H}(X_1 = c+1, \dots, X_M) + \mathcal{U}(v)\big) \\
&\quad\quad -\mathrm{ReLU}\big(\beta + \alpha_1 c + \cdots + \alpha_M X_M + \mathcal{S}(X_1 = c, \dots, X_M) + \mathcal{H}(X_1 = c, \dots, X_M) + \mathcal{U}(v)\big)
\end{aligned} \tag{13}$$

$\mathcal{U}(v)$ is non-negative. Therefore, the precise total effect of $X_1$ is

$$\begin{aligned}
\Delta\widehat{ADV}(X_1 = c) &= \Delta\mathbb{E}_{X_2,\dots,X_M}\widehat{ADV}(X_1 = c, X_2, \dots, X_M) \\
&= \Delta\int \dots \int_{X_2,\dots,X_M} \widehat{ADV}(X_1 = c, X_2, \dots, X_M) p(X_1 = c, X_2, \dots, X_M)\, d(X_2) \dots d(X_M)
\end{aligned} \tag{14}$$

Equation 14 is intractable because of the integral computation. In order to facilitate such computation, we utilize the Monte Carlo method. Equation 14 can be transformed to:

$$\frac{1}{K}\sum_{k=1}^{K} \widehat{ADV}\big(x_{k,1} = c, x_{k,2}, \dots, x_{k,M}\big) \tag{15}$$

where $\big(x_{k,1} = c, x_{k,2}, \dots, x_{k,M}\big)$ denotes the $k$-th sample drawn from the distribution $p(X_1 = c, X_2, \dots, X_M)$. In order to compute the precise total effect of $X_1$, it is necessary to learn the distribution $p(X_1 = c, X_2, \dots, X_M)$, so that samples can be drawn from it. $X_1, X_2, \dots, X_M$ are very sparse in the Euclidean space when using Monte Carlo method. In order to learn a smooth and accurate distribution, we embody a generative adversarial network (GAN) to learn $p(X_1 = c, X_2, \dots, X_M)$. To overcome the instability issues of GANs, we leverage the Wasserstein GAN with gradient penalty (WGAN-GP). We cohesively embed WGAN-GP in our model. The learning loss of the discriminator in our proposed method is given by

$$L_{\mathrm{d}} = \mathbb{E}_{\widetilde{\boldsymbol{x}}\sim\mathbb{P}_{\mathrm{g}}}[\mathrm{D}(\widetilde{\boldsymbol{x}})] - \mathbb{E}_{\boldsymbol{x}\sim\mathbb{P}_{\mathrm{r}}}[\mathrm{D}(\boldsymbol{x})] + \lambda\mathbb{E}_{\widehat{\boldsymbol{x}}\sim\mathbb{P}_{\widehat{x}}}[(\|\nabla_{\widehat{x}}\mathrm{D}(\widehat{\boldsymbol{x}})\|_2 - 1)^2] \tag{16}$$

where $\mathrm{D}(\cdot)$ is a score that measures the quality of the input sample. $\mathbb{P}_{\mathrm{r}}$ is the real distribution. $\mathbb{P}_{\mathrm{g}}$ is the learned distribution by the generator. $\hat{x}$ is sampled uniformly along the straight lines between pairs of points sampled from $\mathbb{P}_{\mathrm{r}}$ and $\mathbb{P}_{\mathrm{g}}$. The distribution of $\hat{x}$ is denoted as $\mathbb{P}_{\hat{x}}$. $\mathbb{E}_{\hat{x}\sim\mathbb{P}_{\hat{x}}}[(\|\nabla_{\hat{x}}\mathrm{D}(\widehat{\boldsymbol{x}})\|_2 - 1)^2]$ is the gradient penalty. $\lambda$ is a positive scalar to control the degree of

the penalty. The loss of the generator is:

$$L_{\mathrm{g}} = -\mathbb{E}_{x\sim\mathbb{P}_{\mathrm{g}}}[\mathrm{D}(\tilde{\boldsymbol{x}})] \tag{17}$$

The trained generator can closely approximate the real distribution $\mathbb{P}_{\mathrm{r}}$, which is fed to Equations 12-17 to yield the precise total effects. Improving upon W&Ds, our approach corrects the feature effect from $\alpha_1$ to $\Delta \int \dots \int_{X_2,\dots,X_M} \widehat{ADV}(X_1 = c, X_2, \dots, X_M) p(X_1 = c, X_2, \dots, X_M)\, d(X_2) \dots d(X_M)$. Such a difference is reflected in different feature rankings and weights, whose material impact on the interpretation is shown in the empirical analyses. Appendix 2 shows the PrecWD algorithm. Figure 1 shows its architecture. [Insert Figure 1 here]

*Novelty of PrecWD*

PrecWD has two original and novel elements: 1) The previous subsection proposes a novel interpretation component that differentiates from W&D. W&D approximates the total effect using the main effect, while our interpretation component is able to offer a precise total effect for the prediction using Equations 12-17. In order to capture the dynamic total effect for each feature, our model predicts the total effect at every feature value. 2) We also design an unstructured component that extends the applicability of W&D to unstructured data analytics.

## Empirical Analyses

### ***Case Study 1: Health Video Viewership Prediction***

*Data Preparation*

Due to the societal impact of healthcare and the timeliness of COVID-19, we first examine the utility of PrecWD in health video viewership prediction. We collected videos from well-known health organizations' YouTube channels, including NIH, CDC, WHO, FDA, Mayo Clinic, Harvard Medicine, Johns Hopkins Medicine, MD Anderson, and Jama. We generated a dataset of 6,528 videos (298 GB). From the data perspective, this study falls into the category of unstructured analytics of viewership prediction, as we directly feed unstructured raw videos as well as video-based features into our model, in addition to webpage metadata. These video-based features and raw videos are directly relevant to video shooting and editing, articulated in Appendix 3. This is in contrast with most existing viewership prediction studies that only use structured webpage metadata as features, such as duration, resolution, dimension, title, and channel ID, among others [42], which lack actionable instructions for video production. Our data size is large among unstructured predictive analytics studies [65,72] and significantly larger than common video-based deep-learning analytics benchmarking datasets, which range from 66 to 4,000 videos [17].

The raw videos are directly fed into PrecWD via the unstructured component. We also generate the commonly adopted video features using BRISQUE. We utilize Liborosa to compute the acoustic features. In order to generate transcripts, we develop a speech recognition model based on DeepSpeech that achieves a 7.06% word error rate on the LibriSpeech corpus. The description, webpage, and channel features are extracted from the webpage. A description of all the features and practical actions on feature interpretation are available in Appendix 3. This case study presents the most prominent features, but our model is not confined to these features. It is a generalized precise interpretable model that can take other features as needed by the end users.

*Evaluation of Predictive Accuracy*

Based on the viewership prediction and interpretable ML literature, we design two groups of baselines: black-box (ML and deep learning) [20,64,66,67,76,77,78] and interpretable methods [13,28,61,70]. The configurations of these baselines are reported in Appendix 4. For all the following analyses, we adopt 10-fold cross-validation where the dataset is divided into 10 folds. Each time we use one fold for test, one fold for validation, and eight folds for training. All the performances in the empirical analysis are the average performance of 10-fold cross-validation. Our model converged as evidenced in Table 2. Table 3 shows the prediction comparison with black-box methods. [Insert Tables 2 and 3 here]

Following recent interpretable ML studies (reviewed in Appendix 5), since this study focuses on providing a new interpretation mechanism, our goal for the predictive accuracy comparison is to be at least on par with, if not better than, the best-performing black-box benchmarks, so that our prediction is reliable and trustworthy. Compared with the best ML method KNN-3, PrecWD reduces MSE by 17.815 ($p < 0.001$). Compared with the best deep learning method LSTM-2, PrecWD reduces MSE by 26.467 ($p < 0.001$). PrecWD remains the best when we fine-tune the benchmarks. These results suggest our prediction is reliable, even though the prediction improvement is not our primary contribution. The main downside of these black-box methods is that they cannot offer a feature-based interpretation, which is critical from the perspectives of trust, model adoption, regulatory enforcement, algorithm transparency, and practical implications and interventions for stakeholders. Extending the line of interpretability, we compare PrecWD with the state-of-the-art interpretable methods in Table 4. Compared with the best interpretable method W&D, PrecWD reduces MSE by 35.072 ($p < 0.001$). [Insert Table 4 here]

Such enhanced predictive accuracy has a significant practical value. Content creators can use our model to help with promotional budget allocation. According to fiverr.com and veefly.com, increasing one view costs an average of $0.03. We perform a detailed benefit analysis in Appendix 6, which suggests that the average annual benefit of our prediction enhancement over the baseline is up to $6.57 billion, and the unreached viewership is reduced by up to 73 billion views. Sponsors can use our model to understand the viewership of the sponsored videos. If content creators opt to pay for promotion to meet sponsors' expectations, such cost will ultimately be transferred to the sponsorship cost, and the economic analysis is similar to that for the content creators.

We fine-tune the hyperparameters of PrecWD in Table 5 to search for the best predictive performance. The hyperparameters include the number of hidden layers and the number of neurons in each layer. We also replace the higher-order component in PrecWD with other deep neural networks, including CNN, LSTM, and BLSTM, to evaluate our design choice. The final model has 3 dense layers in the higher-order component and 16 neurons in each layer. To ensure fair comparisons, all the baseline methods in Tables 3-4 underwent the same parameter-tuning process, and we reported the final fine-tuned results above. [Insert Table 5 here]

We further perform ablation studies to test the efficacy of the individual components of PrecWD. The upper part of Table 6 shows that removing any component of PrecWD negatively impacts the performance, suggesting optimal design choices. In order to test the effectiveness of each feature group, we remove each feature group stepwise and test its contribution to the performance. The lower part of Table 6 shows that removing any feature group will hamper the performance. [Insert Table 6 here]

*Interpretation of PrecWD*

As our core contribution, PrecWD can offer precise total effect using the proposed interpretation component. The WGAN-GP layer in the interpretation component generates samples to learn the data distribution to facilitate the computation of Equations 12-17. While the standard GAN might have a convergence problem, WGAN-GP resolves this issue by penalizing the norm of the gradient of the discriminator with respect to its input. Figure 2.a. shows that the generator loss and discriminator loss both converge in this study. Figure 2.b. shows the discriminator validation loss shrinks together with the training loss and both converge, suggesting the discriminator does not overfit in this training process. We then evaluate the quality of the generated samples. Figure 3.a shows the real samples and the generated samples are inseparable, suggesting the generated samples follow a distribution similar to the real ones. In addition, we use Principal Component Analysis to reduce the feature dimensions to 10 dimensions. Table 7 suggests that WGAN-GP can accurately generate samples whose distribution has no statistical difference from the real samples. We also examined alternative generative models, including VAE and Bayesian Network, whose generated samples differ from the real samples. This suggests that our generative process is accurate and is the best design choice. We further show the quality of the generated samples by presenting the two most important features in Figure 3.b-c. WGAN-GP is able to generate very accurate distributions of these features. We plot the feature-based interpretations in Figure 4.

[Insert Table 7, Figures 2-4 here]

The transcript and description features have a salient influence on the prediction. These features include medical knowledge, informativeness, readability, and vocabulary richness. The results show that one unit of increase[2] in transcript readability results in an increase of 757.402 predicted average daily views. Medical knowledge, operationalized as the number of medical terms [42], has a sizable influence on the prediction as well. One unit of increase in the transcript medical terms will raise the predicted average daily views by 440.649. This is because an easy-to-read and medically informative transcript or description is associated with higher viewership as the viewers attempt to seek medical information from these videos.

The transcript and description sentiments also significantly affect the prediction. These sentiments in the video bring in personal opinions and experiences, which are relatable to viewers, thus enticing higher viewership. The channel features have a critical influence on the prediction as well. YouTube collects information from verified channels, such as phone numbers. Verified channels signal authenticity and credibility to viewers. Therefore, the viewers are more likely to watch the videos posted by these channels.

PrecWD is also capable of estimating the dynamic total effect. Figure 5 shows three randomly selected examples: description vocabulary richness, description readability, and transcript negative

[2] This unit effect is consistent with the interpretation format of linear regression. Although the prediction capability of linear regression is weak, it offers an easily understandable and largely accepted interpretation mechanism. The weight $\beta_i$ of a variable is usually interpreted as when $X_i$ increases one unit, $y$ will increase $i$. This unit effect format has been commonly adopted in many interpretable machine learning studies for various applications [24]. Readability is the Flesch Reading Ease, formulated as: $206.835 - 1.015\left(\frac{total\ words}{total\ sentences}\right) - 84.6\left(\frac{total\ syllables}{total\ words}\right)$, which is the most popular and the most widely tested and used readability measurement by marketers, research communicators, and policy writers, among many others. Increasing readability means using fewer words in a sentence and using words with fewer syllables.

sentiment. Figure 5.a. shows that the total effect of description vocabulary richness is positive when its value is low. Such a total effect turns negative when description vocabulary richness is high. This is because when description vocabulary richness is low, enriching the vocabulary makes the language more appealing. As the vocabulary richness continues to increase, the description becomes too hard to comprehend and viewers lose interest in the video. Figure 5.b. shows that the total effect of description readability increases when the readability value increases. This could be because when the description is readable, it is also easier for the viewers to understand the medical knowledge and other content in the video. Figure 5.c. indicates that the total effect of transcript negative sentiment increases when the value of transcript negative sentiment increases. When a video is enriched with negative sentiment, it usually contains opinions and commentaries, which may be relatable to the viewers' experience or belief and even entice them to write comments. Those interactions in the comment section further enhance viewership. [Insert Figure 5 here]

*Precise Total Effect v.s. Main Effect*

PrecWD offers the precise interpretation (total effect), while the existing approaches (W&D) could only approximate the interpretation using the main effect. The interpretation error correction by our model has significant improvement on the feature effects. For instance, our model interprets description readability to have a positive influence on viewership. This is because readable descriptions are easy to comprehend, thus attracting viewers. However, the existing approach (main effect) interprets description readability to have a negative influence, contradicting common perception. We also quantify the influence of the interpretation error correction in Table 8, where the total effect and the existing approaches (main effects and regressions) have many opposite signs and very distinct feature importance order. Such differences further validate the contribution of our precise interpretation component. The interpretation errors are the direct reason for mistrust of W&D, which we will show in the user study later. We also perform significance tests of the feature effects in Appendix 7. The vast majority of the feature effects are significant ($p < 0.05$). Only one low-ranked feature is not significant. Nevertheless, its total effect is close to zero as well.

[Insert Table 8 here]

Our precise interpretation has significant practical value. To be paid by YouTube, a content creator needs to obtain 20,000 views, 4,000 watch hours, and 1,000 subscribers. The precise interpretation of PrecWD can help content creators achieve these requirements by deriving three actions, which are inspired by related interpretable ML studies, reviewed in Appendix 8. First, we can sort the features by importance and understand what features are more important in predicting viewership, thus increasing trust from end users. Because monetary incentives are involved in practical usage, content creators, sponsors, and platforms thrive on adopting the most trustworthy model. Second, based on the feature importance order, our model can recommend the most effective features to prioritize for getting higher projected viewership. It also recommends the prioritization order of the work process for content creators when time is limited. Third, content creators can allocate a tiered budget to different features to reach the optimal predicted viewership given the fixed budget. Table 8 suggests when the feature effects are imprecise (baseline), the feature importance order can be significantly off, and the sign of the effect can even be the opposite of common perceptions. Our user study will later validate that the feature importance order and signs of feature effects of our precise interpretation are more trustworthy and helpful than other imprecise interpretations. Once qualified to be paid, content creators can earn money based on the number of views a video receives. They can also receive external sponsorships. YouTube's pay rate hovers between $0.01 and $0.03 per view. PrecWD's interpretation can guide content creators to make adjustments in production and improve viewership for higher returns. Appendix 9 provides a more detailed explanation of the actions that can be derived based on the feature interpretation.

### *Case Study 2: Misinformation Viewership Prediction*

Among all the videos, violative videos, such as misinformation videos, are the most concerning, as they lead viewers to institute ineffective, unsafe, costly, or inappropriate protective measures; undermine public trust in evidence-based messages and interventions; and lead to a range of collateral negative consequences [12]. Early identification and prioritized screening based on potential video popularity are the keys to minimizing undesired broad consequences of misinformation videos. This goal necessitates misinformation viewership prediction as well as the understanding of the factors. Case study 2 evaluates PrecWD by predicting misinformation viewership. A number of trusted sources have identified a set of misinformation videos on YouTube (Appendix 10). We crawled these videos, resulting in 4,445 videos (208 GB of data).

PrecWD achieved consistent leading performance. Table 9 shows that, compared to the best ML model (KNN-3), PrecWD drops MSE by 11.398. Compared with the best deep learning method (CNN-3), PrecWD reduces MSE by 3.988. Compared with the best interpretable model (W&D), PrecWD reduces the MSE by 29.206. Ablation studies (Table 10) show that excluding any component negatively impacts the performance, suggesting good design choices. Table 11 shows the generated data distribution has no difference from the real data distribution. We also performed hyperparameter tuning (Appendix 11). The conclusions are consistent with case study 1 and in favor of our method. [Insert Tables 9-11 here]

The enhanced prediction has profound practical values. According to the website Statista 2020, the worldwide economic loss resulting from misinformation is about 78 billion dollars. With social media being a large part of people's daily life and an important source of information, misinformation shared on these platforms account for a significant portion of the damage. Each view of the videos with health misinformation could lead to a significant burden on the healthcare system and result in negative outcomes for the patients. PrecWD offers a *more accurate popularity* *estimation tool* compared to other prediction models. Using this tool, YouTube is able to more precisely identify potentially popular videos. This helps address the misinformation problem by prioritizing the screening of potentially popular videos, thus minimizing the influence of misinformation in a more accurate manner.

We also interpreted the prediction. The textual features and video sentiments are critical features associated with misinformation viewership. This interpretation sheds light on the management of content quality for video-sharing sites. They could monitor the transcript and description features. For instance, when videos show overwhelmingly negative content, they can be marked for credibility review to prevent the potential wide spread of misinformation. The feature interpretation table and explanations are included in Appendix 11.

***Evaluation of the Precise Interpretation Component (Descriptive Accuracy)***

Since our precise interpretation component is the core contribution, this section compares PrecWD's interpretability with the state-of-the-art interpretable frameworks. There is currently no standard quantitative method to evaluate interpretability. Consequently, most computer science studies only show interpretability without an evaluation [5,10,75]. In the business analytics discipline, a few studies have reached a consensus that conducting user studies via lab or field experiments is the most appropriate approach to evaluate ML interpretability [8,25,37,56]. They design surveys to ask participants to rate the interpretability of a model. Following this practice, we design a user study with five groups in Table 12. We recruited 174 students from two national universities in Asia. They were randomly assigned to one of these five groups. We selected nine control variables, and this study passed randomization checks, where the control variables, summary statistics, and randomization *p*-values are reported in Appendix 12. The full survey can be found in Appendix 13. [Insert Table 12 here]

The participants were assigned an ML model to predict the daily viewership of a YouTube video. We showed them the variables the model uses and the weights of the variables. We disclosed that the more reasonable these variables and weights are, the more accurate the prediction would be, and that their compensation is positively related to the prediction accuracy. To ensure the participants understand how to read the variables and weights, we designed a common training session for all participants. To avoid imposing any bias, the training session is context-free, and the variables are pseudo-coded as variables 1-7. We displayed a pseudo model in the training (Appendix 14). We informed them the weight of a variable indicates its importance. We showed an example: "If the weight of a variable is 0.3, this means increasing this variable by 1 unit, the predicted viewership will increase by 0.3 units." After that, we designed the following two test questions to teach them how to read a model. If the participants choose an incorrect answer, an error message and a hint will appear on the screen (Appendix 15). They need to find the correct answer before proceeding to the next page. This learning process ensures they can understand how to read the interpretation of a model. The pseudo model, test questions, error message, and hint wording are the same for all groups.

1. **Question**: According to the above figure, when using the above model to predict the daily viewership of videos, what are the top two essential variables that have positive effects? (**Options**: Variable 1, Variable 2, Variable 3, Variable 4, Variable 5, Variable 6, Variable 7)
2. **Question**: According to the weights in the figure, if variable 6 increases by 1 unit, how will the above model prediction of video viewership change? (**Options**: Increase by 0.3 unit, Increase by 0.6 unit, Decrease by 0.3 unit, Decrease by 0.6 unit)

We first ask the participants to watch the same YouTube video to familiarize the prediction target and context. This is a health video for tobacco education from WHO. We then show them a screenshot of the video webpage (Appendix 16), depicting the potential variables we can use for prediction. Then, we show them the variables and weights of their assigned model (Figure 6). We choose the seven most important variables by weight, because seven is considered the "Magical Number" in psychology and is the limit of our capacity to process short-term information [45]. To help the participants fully understand Figure 6, we design the following four test questions. If the participants choose an incorrect answer, an error message and a hint will appear on the screen, as shown in Appendix 17. They need to find the correct answer before proceeding to the next page. This learning process aims to teach them to understand the variables and weights of their assigned model. The wordings of the error message, hint, and test questions are the same across all groups.

[Insert Figure 6 here]

1. **Question**: According to the above figure, when using the above model to predict video viewership, please rank the following variables from the most important to the least important. Please put the most important variable on the top and the least important on the bottom. Hint: The importance of a variable can be measured by the weight of the variable. (**Options**: The seven variables in a randomized order)
2. **Question**: What are the top 2 most essential variables in the above model? (**Options**: Four variables)
3. **Question**: If the creator of the above video would like to increase video viewership, increasing/decreasing which variable is more effective? (**Options**: Four variables)
4. **Question**: According to the above figure, if the variable in the bottom row increases by 1 unit, how will the model prediction of video viewership change? (**Options**: Four changes)

After answering these questions correctly, the participants have a good understanding of the assigned model. We then ask them to rate the interpretability of their assigned model. Following literature review, we use two metrics of interpretability: trust in automated systems and model usefulness, adopted from [11] and [4]. The Cronbach's Alpha is 0.963 for the trust in automated systems scale, and 0.975 for the usefulness scale, suggesting excellent reliability. The factor loadings are shown in Appendix 18, showing great validity. We designed an attention check question in the scales ("Please just select neither agree nor disagree"). After removing those who failed the attention check, 140 participants remained. We perform *t*-tests in Table 13 on PrecWD and the baseline groups to compare interpretability. [Insert Table 13 here]

PrecWD has significantly better interpretability than the baseline models. Such improvement is attributed to PrecWD's ability to capture the precise feature effects which lead to the most reasonable and trustworthy variables and ranking. Our model suggests that description readability, medical knowledge, and video sentiment are the most influential variables in predicting the viewership of the given WHO video. These variables are in line with the literature which documented that readability [69], medical knowledge [42], and sentiment [69] are the driving factors for social media readership. Such alignment with prior knowledge and perception gained users' trust. The feature effects in the baseline models, however, are imprecise due to the mismatch between the main effect and the total effect. Such a feature effect error results in many counter-intuitive variables and importance order in the baseline groups. These counter-intuitive variables are the direct reasons for mistrust in the baseline groups. For instance, W&D shows that the appearance of numbers in the video description is the most important variable, which has little to do with the video content. Studies have shown that content characteristics are the leading factors for social media readership [32]. Users may find it difficult to believe the appearance of numbers could predict viewership. SHAP shows more audio tracks reduce viewership, which contradicts common sense, because more audio track options should attract more foreign language viewers. Piecewise W&D and VAE suggest the frequency of two-word phrases is the top variable, while content characteristics, such as medical knowledge and readability, are the least important. Such ranking is the inverse of common understanding and contradicts the literature [32,69]. These counter-intuitive examples significantly reduce users' trust in these models.

After the participants rated interpretability, we conduct a supplementary study to investigate which model they would finally adopt. We inform the participants that, if they think the variables and weights of the previous model are not reasonable, they have a chance to change to a different model. Since we incentivized the participants to choose the most reasonable model, their final adoption indicates the model that they trust the most[3]. Then we show the variables and weights of all the five models, similar to Figure 6. The order of the five models is randomized. We ask which model they would like to finally adopt for the prediction, and we measure the interpretability of the

[3] After the survey, we disclosed how their model performed in relative to the other four models. We compensated them with different-valued office supplies in the end, according to the model performance ranking.

adopted model, reported in Tables 14 and 15. [Insert Tables 14 & 15 here]

105 participants (75%) finally adopted our model. Table 15 shows that, for those who finally adopted our model, the second-time interpretability is higher than the first time. This is because after the participants see all five models, the relative advantage of our model is even more obvious, causing them to rate the interpretability of our model higher the second time.

SHAP's interpretation only shows the feature importance. It cannot reveal the unit-increase effect as our model does. To compare our model and SHAP in the same format, we conduct a second user study with two groups (N = 65) from a university in Asia. For each group, we only show the feature importance without the unit-increase effect (the weight), depicted in Figure 7. Such display shows the variables, their relative importance, and effect direction, in which our model and SHAP are already significantly different. Such a difference sufficiently attributes to interpretability variance. The second user study uses the same control variables and deploys a similar training session, randomization checks, and attention checks, reported in Appendices 19-20. After removing those who failed the attention checks, 55 participants remained. Table 16 suggests our model's interpretability still outperforms SHAP. [Insert Figure 7, Table 16 here]

## Discussion

### *Implications to IS Knowledge Base*

In line with the design science research guidelines [31], this study identifies an impactful problem in social media analytics: video viewership prediction and interpretation. We develop a novel information system that predicts video viewership while interpreting the predictors. We conduct comprehensive evaluations and interpretations of the information system and design two user studies to assess its utility. This study also fits in the computational genre of design science research [51]. Our study develops an interdisciplinary approach that involves a novel computational algorithm and an analytical solution to a major societal problem, thus holding great potential for generating IS research with significant societal impact [3,20,33,41,65,72].

***Implications to Methodology and IS Design Theory***

PrecWD uniquely models unstructured data and proposes a novel interpretation component to offer precise total effect and its dynamic changes. Our model shows two generalizable design principles for model development: 1) Generative models can assist the interpretation of predictive models (based on the model interpretation and the user study result); 2) Raw unstructured data can complement crafted features in prediction (based on Tables 6 and 10). These design principles along with PrecWD provide a "nascent design theory" [27,40] for the design science paradigm of IS studies. Using these design principles, PrecWD can be generalized to understand the underlying factors of predictive analytics in other problem domains.

***Implications to Practice***

This study offers many practical implications. For video-sharing sites, PrecWD is a deployable analytics system that can predict video viewership and offer the interpretation of the prediction. Our model provides insights to monitor video popularity where credible videos can be approved and violative videos can be banned before they are published, thus minimizing widespread infiltration of violative videos. Content creators can leverage our model to predict video viewership in order to determine where to allocate more promotional funds. Our interpretation ensures the trust of the model and gives content creators actionable directions to understand the importance of features and optimize the prioritization of video features in their workflow.

***Limitations and Future Directions***

First, we focus on long-form videos. Future work can design new features related to the song templates of short-form videos. New models can be devised to consider the implicit relationship between short-form video content and background song templates. Second, content creators could pay for promotion to increase video exposure, which cannot be observed publicly. Our features may miss the influence of paid promotions. Nevertheless, this issue is mitigated because most of our videos are collected from non-profit health organizations that publish educational videos. They have little monetary incentive to pay for promotions. Third, recommendations, shares on social networks, long tail, and Mathew effects could influence viewership. Omitting them is another limitation. It is challenging to estimate these effects on a video as they vary significantly among different viewers. We did not attempt to capture these effects in our features. However, the video features reflect the video type, quality, visual and audio effects, and more, which affect whether users would actually watch a recommended video. Channel-level features (e.g., verification) also implicitly characterize the social network, long tail, and Mathew effects, because verified channels are usually more influential. Therefore, our features partially mitigate this limitation. Fourth, since SHAP and model-based methods have different underlying interpretation mechanisms. Quantitatively comparing them in the absolute same presentation is not achievable. User study 2 offers the most approximate comparison design. Nevertheless, *post-hoc* methods like SHAP have been repeatedly recommended against usage because of their unfaithful, vulnerable, and unstable issues. Fifth, as a shared limitation of most interpretable ML models, the interpretation implies correlation not causation. While such correlation can be used to derive many actionable recommendations, these models cannot directly use the feature weights to derive causal actions. To derive causal actions, studies can incorporate econometric models into ML models. Such a causal ML discipline is a largely different yet interesting paradigm for future work.

**Conclusion**

This study proposes PrecWD for viewership prediction and interpretation. To address the pitfalls of prior interpretable frameworks, our study incorporates an unstructured component and innovatively captures the precise total effect as well as its dynamic changes. Empirical results indicate PrecWD outperforms strong baselines. Two user studies confirm that the interpretability of PrecWD is significantly better than other interpretable methods, particularly in improving trust and model usefulness. These findings offer implementable actions for content creators and video-sharing sites to optimize the video production process and manage content quality.

## Acknowledgement

This research was carried out with the support of the "University of Delaware General University Research" fund. Yidong Chai is supported by National Natural Science Foundation of China (72293581, 91846201, 72293580, 72188101).

## References

1. Abbasi, A., Albrecht, C., Vance, A., and Hansen, J. Metafraud: a meta-learning framework for detecting financial fraud. *MIS Quarterly*, *36*, 4 (2012), 1293–1327.
2. Abbasi, A., Zhang, Z., Zimbra, D., Chen, H., and Nunamaker, J.F. Detecting fake websites: the contribution of statistical learning theory. *MIS Quarterly*, *34*, 3 (2010), 435–461.
3. Abbasi, A., Zhou, Y., Deng, S., and Zhang, P. Text analytics to support sense-making in social media: A language-action perspective. *MIS Quarterly*, *42*, 2 (2018), 427–464.
4. Adams, D.A., Nelson, R.R., and Todd, P.A. Perceived usefulness, ease of use, and usage of information technology: A replication. *MIS Quarterly*, *16*, 2 (1992), 227–247.
5. Agarwal, R., Frosst, N., Zhang, X., Caruana, R., and Hinton, G.E. Neural Additive Models: Interpretable Machine Learning with Neural Nets. In *NeurIPS*. 2020, pp. 699–711.
6. Akpinar, E. and Berger, J. Valuable Virality. *Journal of Marketing Research*, *54*, 2 (2017), 318–330.
7. Alqahtani, H., Kavakli-Thorne, M., and Kumar, G. Applications of Generative Adversarial Networks (GANs): An Updated Review. *Archives of Computational Methods in Engineering*, *28*, (2021), 525–552.
8. Bastani, O., Kim, C., and Bastani, H. Interpreting Blackbox Models via Model Extraction. *arXiv*, arXiv preprint arXiv:1705.08504. (2017).
9. Burel, G., Saif, H., and Alani, H. Semantic wide and deep learning for detecting crisis-information categories on social media. In *International Semantic Web Conference*. 2017, pp. 138–155.
10. Caruana, R., Lou, Y., Gehrke, J., Koch, P., Sturm, M., and Elhadad, N. Intelligible Models for HealthCare: Predicting Pneumonia Risk and Hospital 30-day Readmission. *Proceedings of KDD*, (2015), 1721–1730.
11. Chai, S., Das, S., and Rao, H.R. Factors Affecting Bloggers' Knowledge Sharing: An Investigation Across Gender. *Journal of Management Information Systems*, *28*, 3 (2014), 309–342.
12. Chai, Y., Li, W., Zhu, B., Liu, H., and Jiang, Y. An Interpretable Wide and Deep Model for Online Disinformation Detection. *SSRN Electronic Journal*, SSRN 3879632 (2022).
13. Cheng, H.T., Koc, L., Harmsen, J., et al. Wide & deep learning for recommender systems. In *ACM International Conference Proceeding Series*. 2016, pp. 7–10.
14. Cichy, P., Salge, T.O., and Kohli, R. Privacy Concerns And Data Sharing In The Internet Of Things: Mixed Methods Evidence From Connected Cars. *MIS Quarterly*, *45*, 4 (2021), 1863–1891.
15. Croitoru, F.-A., Hondru, V., Ionescu, R.T., and Shah, M. Diffusion Models in Vision: A Survey. *IEEE TRANSACTIONS ON PATTERN ANALYSIS AND MACHINE INTELLIGENCE*, *14*, 8 (2022), 1–22.
16. Dhurandhar, A., Iyengar, V., Luss, R., and Shanmugam, K. TIP: Typifying the Interpretability of Procedures. *arXiv* , preprint arXiv:1706.02952. (June 2017).
17. Dong, L.I.U., Yue, L.I., Jianping, L.I.N., Houqiang, L.I., and Feng, W.U. Deep Learning-Based Video Coding: A Review and A Case Study. *ACM Computing Surveys*, *53*, 1 (2019), 1–35.
18. Duan, W., Gu, B., and Whinston, A.B. The dynamics of online word-of-mouth and product sales—An empirical investigation of the movie industry. *Journal of Retailing*, *84*, 2 (2008), 233–242.
19. Ebrahimi, M., Chai, Y., Samtani, S., and Chen, H. Cross-Lingual Cybersecurity Analytics in the International Dark Web with Adversarial Deep Representation Learning. *MIS Quarterly*, *Forthcoming*, (2022).
20. Ebrahimi, M., Nunamaker, J.F., and Chen, H. Semi-Supervised Cyber Threat Identification in Dark Net Markets:

A Transductive and Deep Learning Approach. *Journal of Management Information Systems*, *37*, 3 (2020), 694–722.
21. Fang, X., Hu, P.J.-H.H., Li, Z. (Lionel) L., and Tsai, W. Predicting adoption probabilities in social networks. *Information Systems Research*, *24*, 1 (2013), 128–145.
22. Ferguson, R. Word of mouth and viral marketing: Taking the temperature of the hottest trends in marketing. *Journal of Consumer Marketing*, *25*, 3 (2008), 179–182.
23. Fortuna, G., Schiavo, J.H., Aria, M., Mignogna, M.D., and Klasser, G.D. The usefulness of YouTube™ videos as a source of information on burning mouth syndrome. *Journal of oral rehabilitation*, *46*, 7 (2019), 657–665.
24. Garnica-Caparrós, M. and Memmert, D. Understanding gender differences in professional European football through machine learning interpretability and match actions data. *Scientific Reports*, *11*, 1 (2021), 1–14.
25. Goldstein Jake M Hofman, D.G., Wortman Vaughan, J., Poursabzi-Sangdeh, F., Goldstein, D.G., Hofman, J.M., and Wallach, H. Manipulating and measuring model interpretability. In *Conference on Human Factors in Computing Systems*. 2021, pp. 1–52.
26. Goyal, Y., Feder, A., Shalit, U., and Kim, B. Explaining Classifiers with Causal Concept Effect (CaCE). *arXiv*, preprint arXiv:1907.07165 (July 2019).
27. Gregor, S. and Hevner, A. Positioning and Presenting Design Science Research for Maximum Impact. *MIS Quarterly*, *37*, 2 (2013), 337–355.
28. Guo, M., Zhang, Q., Liao, X., and Zeng, D.D. An interpretable neural network model through piecewise linear approximation. *arXiv*, preprint arXiv:2001.07119 (2020).
29. Han, Y., Chen, W., Xiong, X., Li, Q., Qiu, Z., and Wang, T. Wide & Deep Learning for improving Named Entity Recognition via Text-Aware Named Entity Normalization. In *Thirty-Third AAAI Conference on Artificial Intelligence*. 2019.
30. Heider, F. and Simmel, M. An Experimental Study of Apparent Behavior. *The American Journal of Psychology*, *57*, 2 (1944), 243–259.
31. Hevner, A.R., March, S.T., Park, J., and Ram, S. Design Science in Information Systems Research. *MIS Quarterly*, *28*, 1 (2004), 75–105.
32. Jaakonmäki, R., Müller, O., and vom Brocke, J. The impact of content, context, and creator on user engagement in social media marketing. In *HICSS*. 2017, pp. 1152–1160.
33. Karahanna, E., Xu, S.X., Xu, Y., and Zhang, N. The needs-affordances-features perspective for the use of social media. *MIS Quarterly*, *42*, 3 (2018), 737–756.
34. Koch, C., Werner, S., Rizk, A., and Steinmetz, R. MIRA: Proactive music video caching using convnet-based classification and multivariate popularity prediction. In *26th IEEE International Symposium on Modeling, Analysis and Simulation of Computer and Telecommunication Systems*. 2018, pp. 109–115.
35. Krijestorac, H., Garg, R., and Mahajan, V. Cross-Platform Spillover Effects in Consumption of Viral Content: A Quasi-Experimental Analysis Using Synthetic Controls. *Information Systems Research*, *31*, 2 (2020), 449–472.
36. Laugel, T., Lesot, M.-J., Marsala, C., Renard, X., and Detyniecki, M. The Dangers of Post-hoc Interpretability: Unjustified Counterfactual Explanations. In *Proceedings of the Twenty-Eighth International Joint Conference on Artificial Intelligence (IJCAI-19)*. 2019, pp. 2801–2807.
37. Lee, D., Manzoor, E., and Cheng, Z. Focused Concept Miner (FCM): Interpretable Deep Learning for Text Exploration. *SSRN Electronic Journal*, SSRN 3304756 (May 2018).
38. Lee, J.W. and Chan, Y.Y. Fine-Grained Plant Identification using wide and deep learning model. In *2019 International Conference on Platform Technology and Service*. 2019, pp. 1–5.
39. Li, J., Larsen, K., and Abbasi, A. Theoryon: A Design Framework And System For Unlocking Behavioral Knowledge Through Ontology Learning. *MIS Quarterly*, *44*, 4 (2020), 1733–1772.
40. Lin, Y.K., Chen, H., Brown, R.A., Li, S.H., and Yang, H.J. Healthcare predictive analytics for risk profiling in chronic care. *MIS Quarterly*, *41*, 2 (2017), 473–495.
41. Lin, Y.-K. and Fang, X. First, Do No Harm: Predictive Analytics to Reduce In-Hospital Adverse Events. *Journal of Management Information Systems*, *38*, 4 (2021), 1122–1149.
42. Liu, X., Zhang, B., Susarla, A., and Padman, R. Go to YouTube and Call Me in the Morning: Use of Social Media for Chronic Conditions. *MIS Quarterly*, *44*, 1 (2020), 257–283.
43. Lundberg, S.M. and Lee, S.-I. A Unified Approach to Interpreting Model Predictions. In *Advances in Neural Information Processing Systems 2017*. 2017.
44. Mai, F., Shan, Z., Bai, Q., Wang, X. (Shane), and Chiang, R.H.L. How Does Social Media Impact Bitcoin Value? A Test of the Silent Majority Hypothesis. *Journal of Management Information Systems*, *35*, 1 (2018), 19–52.
45. Miller, G.A. The magical number seven, plus or minus two: some limits on our capacity for processing information. *Psychological Review*, *63*, 2 (1956), 81–97.
46. Miller, T. Explanation in artificial intelligence: Insights from the social sciences. *Artificial Intelligence*, *267*,

(2019), 1–38.
47. Molnar, C. *Interpretable Machine Learning*. Leanpub, 2019.
48. Montavon, G., Binder, A., Lapuschkin, S., Samek, W., and Müller, K.-R. Layer-Wise Relevance Propagation: An Overview. *Lecture Notes in Computer Science*, (2019), 193–209.
49. Murdoch, W.J., Singh, C., Kumbier, K., Abbasi-Asl, R., and Yu, B. Definitions, methods, and applications in interpretable machine learning. *Proceedings of the National Academy of Sciences of the United States of America*, *116*, 44 (2019), 22071–22080.
50. NYTimes. YouTube Discloses Percentage of Views That Go to Videos That Break its Rules - The New York Times. 2021. https://www.nytimes.com/2021/04/06/technology/youtube-views.html.
51. Rai, A. Editor's comments: diversity of Design Science Research. *MIS Quarterly*, *41*, 1 (2017), iii–xviii.
52. Ribeiro, M.T., Singh, S., and Guestrin, C. "Why should i trust you?" Explaining the predictions of any classifier. *Proceedings of the ACM SIGKDD International Conference on Knowledge Discovery and Data Mining*, (2016), 1135–1144.
53. Ruby, D. YouTube Statistics (2022) — Updated Data, Facts & Figures Shared! *DEMANDSAGE*, 2022. https://www.demandsage.com/youtube-stats/.
54. Saboo, A.R. Using Big Data to Model Time-Varying Effects for Marketing Resource (RE) Allocation. *MIS Quarterly*, *40*, 4 (2016), 911–939.
55. Samtani, S., Chai, Y., and Chen, H. Linking exploits from the dark web to known vulnerabilities for proactive cyber threat intelligence: An attention-based deep structured semantic model. *MIS Quarterly*, *Forthcoming*, (2021).
56. Selvaraju, R.R., Cogswell, M., Das, A., Vedantam, R., Parikh, D., and Batra, D. Grad-CAM: Visual Explanations From Deep Networks via Gradient-Based Localization. 2017, 618–626.
57. Shin, D., He, S., Lee, G.M., Whinston, A.B., Cetintas, S., and Lee, K.-C. Enhancing Social Media Analysis with Visual Data Analytics: A Deep Learning Approach. *MIS Quarterly*, *44*, 4 (2020), 1459–1492.
58. Siegmund, N., Kolesnikov, S.S., Kästner, C., et al. Predicting performance via automated feature-interaction detection. In *International Conference on Software Engineering*. 2012, pp. 167–177.
59. Slack, D., Hilgard, S., Jia, E., Singh, S., and Lakkaraju, H. Fooling LIME and SHAP: Adversarial Attacks on Post hoc Explanation Methods. In *Proceedings of the AAAI Conference on AI, Ethics, and Society*. 2020, pp. 180–186.
60. Stieglitz, S. and Dang-Xuan, L. Emotions and information diffusion in social media - Sentiment of microblogs and sharing behavior. *Journal of Management Information Systems*, *29*, 4 (2013), 217–248.
61. Tosun, N., Sert, E., Ayaz, E., Yilmaz, E., and Gol, M. Solar Power Generation Analysis and Forecasting Real-World Data Using LSTM and Autoregressive CNN. In *In 2020 International Conference on Smart Energy Systems and Technologies*. 2020, pp. 1–6.
62. Tsang, M., Cheng, D., and Liu, Y. Detecting Statistical Interactions from Neural Network Weights. In *ICLR 2018*. 2018.
63. Vynck, G. de and Lerman, R. Facebook and YouTube are still full of covid misinformation - The Washington Post. *The Washington Post*, 2021. https://www.washingtonpost.com/technology/2021/07/22/facebook-youtube-vaccine-misinformation/.
64. Xie, J., Liu, X., Zeng, D., and Fang, X. Understanding Reasons for Medication Nonadherence: An Exploration in Social Media Using Sentiment-Enriched Deep Learning Approach. In *ICIS 2017 Proceedings*. 2017.
65. Xie, J., Liu, X., Zeng, D.D., and Fang, X. Understanding Medication Nonadherence from Social Media: A Sentiment-Enriched Deep Learning Approach. *MIS Quarterly*, *46*, 1 (2022), 341–372.
66. Xie, J. and Zhang, B. Readmission Risk Prediction for Patients with Heterogeneous Hazard: A Trajectory-Aware Deep Learning Approach. In *ICIS 2018 Proceedings*. 2018.
67. Xie, J., Zhang, Z., Liu, X., and Zeng, D. Unveiling the Hidden Truth of Drug Addiction: A Social Media Approach Using Similarity Network-Based Deep Learning. *Journal of Management Information Systems*, *38*, 1 (2021), 166–195.
68. Xie, L., Hu, Z., Cai, X., Zhang, W., and Chen, J. Explainable recommendation based on knowledge graph and multi-objective optimization. *Complex & Intelligent Systems 2021 7:3*, *7*, 3 (2021), 1241–1252.
69. Yang, M., Ren, Y., and Adomavicius, G. Understanding User-Generated Content and Customer Engagement on Facebook Business Pages. *Information Systems Research*, *30*, 3 (2019), 839–855.
70. Ye, H., Cao, B., Peng, Z., Chen, T., Wen, Y., and Liu, J. Web Services Classification Based on Wide & Bi-LSTM Model. *IEEE Access*, *7*, (2019), 43697–43706.
71. Yu, H., Xie, L., and Sanner, S. The Lifecyle of a Youtube Video: Phases, Content and Popularity. In *International AAAI Conference on Web and Social Media*. 2015, pp. 533–542.
72. Yu, S., Chai, Y., Chen, H., Brown, R.A., Sherman, S.J., and Nunamaker, J.F. Fall Detection with Wearable Sensors: A Hierarchical Attention-based Convolutional Neural Network Approach. *Journal of Management*

*Information Systems*, *38*, 4 (2021), 1095–1121.

73. Yu, S., Chai, Y., Chen, H., Sherman, S.J., and Brown, R.A. Wearable Sensor-based Chronic Condition Severity Assessment: An Adversarial Attention-based Deep Multisource Multitask Learning Approach. *MIS Quarterly*, *Forthcoming*, (2022).
74. Zafar, M.R. and Khan, N. Deterministic Local Interpretable Model-Agnostic Explanations for Stable Explainability. *Machine Learning and Knowledge Extraction*, *3*, 3 (2021), 525–541.
75. Zeiler, M.D. and Fergus, R. Visualizing and Understanding Convolutional Networks. *Lecture Notes in Computer Science*, (2014), 818–833.
76. Zhang, D., Zhou, L., Kehoe, J.L., and Kilic, I.Y. What Online Reviewer Behaviors Really Matter? Effects of Verbal and Nonverbal Behaviors on Detection of Fake Online Reviews. *Journal of Management Information Systems*, *33*, 2 (April 2016), 456–481.
77. Zhu, H., Samtani, S., Brown, R.A., and Chen, H. A deep learning approach for recognizing activity of daily living (adl) for senior care: Exploiting interaction dependency and temporal patterns. *MIS Quarterly*, *45*, 2 (2021), 859–896.
78. Zhu, H., Samtani, S., Chen, H., and Nunamaker, J.F. Human Identification for Activities of Daily Living: A Deep Transfer Learning Approach. *Journal of Management Information Systems*, *37*, 2 (2020), 457–483.

## Tables

**Table 1. Recent Work in Interpretable Machine Learning Methods**

| Model | Scope | Data | Method | Usage | Model | Scope | Data | Method | Usage |
|---|---|---|---|---|---|---|---|---|---|
| SHAP [43] | Both | Any | Perturbation | Post-hoc | GAM [10] | Model-level | Tabular | GAM | Model-based |
| LIME [52] | Both | Any | Perturbation | Post-hoc | W&D [13] | Model-level | Tabular | W&D | Model-based |
| LRP [48] | Both | Any | Backprop | Post-hoc | W&D-CNN [9] | Model-level | Tabular | W&D | Model-based |
| Deconvolutional Nets [75] | Instance-level | Image | Backprop | Post-hoc | W&D-LSTM [61] | Model-level | Tabular | GAM | Model-based |
| CaCE [26] | Model-level | Image | GAM | Post-hoc | W&D-BLSTM [70] | Model-level | Tabular | W&D | Model-based |
| Grad-CAM [56] | Instance-level | Image | Backprop | Post-hoc | Piecewise W&D [28] | Model-level | Tabular | W&D | Model-based |
| NAM [5] | Model-level | Any | GAM | Model-based | Our method | Both | Any | PrecWD | Model-based |

**Table 2. Training and Test Loss**

| Epoch | 1 | 2 | 3 | 4 | 5 | 6 | 7 | 8 | 9 | 10 | 11 | 12 | 13 | 14 | 15 | 16 | 17 | 18 | 19 | 20 | 21 | 22 | 23 | 24 |
|---|---|---|---|---|---|---|---|---|---|---|---|---|---|---|---|---|---|---|---|---|---|---|---|---|
| Training | 3.57 | 2.62 | 2.27 | 2.22 | 2.19 | 2.18 | 2.16 | 2.15 | 2.14 | 2.13 | 2.13 | 2.12 | 2.12 | 2.11 | 2.09 | 2.08 | 2.07 | 2.06 | 2.06 | 2.05 | 2.04 | 2.04 | 2.04 | 2.04 |
| Test | 3.07 | 2.21 | 2.20 | 2.19 | 2.18 | 2.18 | 2.17 | 2.17 | 2.17 | 2.16 | 2.16 | 2.17 | 2.15 | 2.15 | 2.15 | 2.15 | 2.16 | 2.15 | 2.15 | 2.16 | 2.16 | 2.16 | 2.16 | 2.16 |

**Table 3. Prediction Comparison of PrecWD with Black-box Methods**

| Method | Outcome Variable: ADV | | Outcome Variable: log(total view) (add published_days as a feature) | | Outcome Variable: ADV (add published_days as a feature) | |
|---|---|---|---|---|---|---|
| | **MSE** | **MSLE** | **MSE** | **MSLE** | **MSE** | **MSLE** |
| **PrecWD (Ours)** | **165.442** | **0.992** | **2.411** | **0.031** | **165.296** | **0.991** |
| Linear regression | 285.585*** | 2.924*** | 2.520** | 0.044** | 288.174*** | 2.931*** |
| KNN-1 | 196.114*** | 2.961*** | 3.668*** | 0.058*** | 192.582*** | 2.940*** |
| KNN-3 | 183.257*** | 2.994*** | 2.616*** | 0.042** | 182.735*** | 2.990*** |
| KNN-5 | 398.577*** | 4.294*** | 2.507*** | 0.039** | 370.482*** | 3.928*** |
| DT-MSE | 382.407*** | 3.624*** | 4.062*** | 0.066*** | 378.914*** | 3.615*** |
| DT-MAE | 374.503*** | 4.065*** | 5.071*** | 0.079*** | 377.583*** | 4.017*** |
| DT-Fredmanmse | 216.523*** | 2.701*** | 4.438*** | 0.069*** | 214.485*** | 2.700*** |
| SVR-Linear | 185.616*** | 4.726*** | 3.553*** | 0.054*** | 185.583*** | 4.722*** |
| SVR-RBF | 219.136*** | 2.836*** | 3.540*** | 0.054*** | 210.105*** | 2.807*** |
| SVR-Poly | 201.043*** | 3.658*** | 3.575*** | 0.054*** | 197.530*** | 3.644*** |
| SVR-Sigmoid | 398.577*** | 4.294*** | 3.571*** | 0.054*** | 391.519*** | 4.287*** |
| Gaussian Process-1 | 999.058*** | 20.587*** | 4.172*** | 1.046*** | 997.833*** | 20.591*** |
| Gaussian Process-3 | 997.295*** | 20.222*** | 4.347*** | 1.036*** | 995.534*** | 20.185*** |
| Gaussian Process-5 | 995.329*** | 19.890*** | 4.354*** | 1.036*** | 995.290*** | 20.013*** |
| MLP-1 | 364.064*** | 3.094*** | 2.668*** | 0.038* | 366.852*** | 3.152*** |
| MLP-2 | 351.437*** | 2.852*** | 2.635** | 0.039* | 353.429*** | 2.862*** |
| MLP-3 | 281.090*** | 2.728*** | 2.721*** | 0.042* | 280.441*** | 2.705*** |
| MLP-4 | 279.648*** | 2.550*** | 2.663*** | 0.035* | 277.118*** | 2.503*** |
| CNN-1 | 207.271*** | 1.453*** | 2.785*** | 0.037* | 203.282*** | 1.417*** |
| CNN-2 | 193.240*** | 1.333*** | 2.687*** | 0.039* | 196.350*** | 1.352*** |
| CNN-3 | 199.997*** | 1.414*** | 2.824*** | 0.039* | 199.824*** | 1.419*** |
| CNN-4 | 196.158*** | 1.326*** | 2.790*** | 0.040* | 198.525*** | 1.332*** |
| LSTM-1 | 271.600*** | 1.692*** | 2.850*** | 0.039* | 275.219*** | 1.728*** |
| LSTM-2 | 191.909*** | 1.185*** | 2.819*** | 0.038* | 193.542*** | 1.194*** |
| BLSTM-1 | 354.760*** | 1.942*** | 2.724*** | 0.037* | 342.729*** | 1.867*** |
| BLSTM-2 | 196.247*** | 1.212*** | 2.947*** | 0.039** | 193.573*** | 1.202*** |

*: $p < 0.05$; **: $p < 0.01$; ***: $p < 0.001$; The details of the baseline models are reported in Appendix 4.

## Table 4. Prediction Comparison of PrecWD with Interpretable Deep Learning

| Method | Outcome Variable: ADV | | Outcome Variable: log(total view) (add published_days as a feature) | | Outcome Variable: ADV (add published_days as a feature) | |
|---|---|---|---|---|---|---|
| | **MSE** | **MSLE** | **MSE** | **MSLE** | **MSE** | **MSLE** |
| **PrecWD (Ours)** | **165.442** | **0.992** | **2.411** | **0.031** | **165.296** | **0.991** |
| W&D [13] | 200.514*** | 1.300*** | 2.560** | 0.036* | 198.538*** | 1.285*** |
| W&D-CNN [38] | 222.786*** | 1.373*** | 2.791*** | 0.038* | 218.849*** | 1.364*** |
| W&D-LSTM [61] | 217.988*** | 1.343*** | 2.566** | 0.037* | 219.533*** | 1.351*** |
| W&D-BLSTM [70] | 208.775*** | 1.233*** | 2.559** | 0.035* | 208.714*** | 1.230*** |
| Piecewise W&D-10 [28] | 268.205*** | 1.696*** | 2.807*** | 0.039* | 270.180*** | 1.699*** |
| Piecewise W&D-20 [28] | 231.900*** | 1.390*** | 2.950*** | 0.037* | 230.581*** | 1.384*** |

*: $p < 0.05$; **: $p < 0.01$; ***: $p < 0.001$; Piecewise W&D-$i$: divide the input into $i$ intervals; The details of the baseline models are reported in Appendix 4

## Table 5. Hyperparameter Fine-tuning

| Method-Network-Layer-Neuron | MSE | MSLE | Method-Network-Layer-Neuron | MSE | MSLE | Method-Network-Layer-Neuron | MSE | MSLE | Method-Network-Layer-Neuron | MSE | MSLE |
|---|---|---|---|---|---|---|---|---|---|---|---|
| PrecWD-Dense-1-16 | 168.004 | 1.015 | PrecWD-Dense-3-32 | 167.930 | 1.010 | PrecWD-CNN-2-64 | 170.532 | 1.080 | PrecWD-LSTM-2-16 | 168.650 | 0.984 |
| PrecWD-Dense-1-32 | 169.325 | 1.034 | PrecWD-Dense-3-64 | 166.352 | 1.040 | PrecWD-CNN-3-32 | 174.986 | 1.132 | PrecWD-LSTM-2-32 | 171.583 | 0.985 |
| PrecWD-Dense-1-64 | 171.535 | 1.053 | PrecWD-CNN-1-16 | 177.644 | 1.270 | PrecWD-CNN-2-16 | 178.285 | 1.143 | PrecWD-BLSTM-1-16 | 170.128 | 0.996 |
| PrecWD-Dense-2-16 | 168.033 | 1.023 | PrecWD-CNN-1-32 | 178.385 | 1.303 | PrecWD-CNN-3-64 | 179.953 | 1.154 | PrecWD-BLSTM-1-32 | 173.250 | 0.994 |
| PrecWD-Dense-2-32 | 168.734 | 1.016 | PrecWD-CNN-1-64 | 180.299 | 1.320 | PrecWD-LSTM-1-16 | 175.494 | 1.015 | PrecWD-BLSTM-2-16 | 169.666 | 0.992 |
| PrecWD-Dense-2-64 | 170.598 | 1.077 | PrecWD-CNN-2-32 | 169.539 | 1.054 | PrecWD-LSTM-1-32 | 170.644 | 0.995 | PrecWD-BLSTM-2-32 | 169.232 | 0.987 |
| **PrecWD-Dense-3-16** | **165.442** | **0.992** | | | | | | | | | |

## Table 6. Ablation Studies

| Method | Outcome Variable: ADV | | Outcome Variable: log(total view) (add published_days as a feature) | | Outcome Variable: ADV (add published_days as a feature) | |
|---|---|---|---|---|---|---|
| | **MSE** | **MSLE** | **MSE** | **MSLE** | **MSE** | **MSLE** |
| **PrecWD** | **165.442** | **0.992** | **2.411** | **0.031** | **165.296** | **0.991** |
| PrecWD without Unstructured Component | 181.056** | 1.060* | 2.523** | 0.037* | 180.029** | 1.057* |
| PrecWD without Piecewise Linear Component | 213.434*** | 1.340*** | 3.445*** | 0.042** | 215.283*** | 1.352*** |
| PrecWD without Second-Order Component | 187.401*** | 1.107** | 2.554** | 0.037* | 188.977*** | 1.164** |
| PrecWD without High-order Component | 191.891*** | 1.152** | 2.584*** | 0.038* | 189.538*** | 1.146** |
| PrecWD with Simple Linear Encoding | 189.712*** | 1.063* | 2.491* | 0.037* | 189.281*** | 1.061* |
| PrecWD with 10 Ordinal One-hot Encoding | 190.224*** | 1.047* | 2.523** | 0.038* | 188.433*** | 1.044* |
| PrecWD with 20 Ordinal One-hot Encoding | 186.767** | 0.994 | 2.483* | 0.037* | 185.251** | 0.992 |
| PrecWD with 10 Ordinal Encoding | 191.969*** | 1.149*** | 2.477* | 0.036* | 191.025*** | 1.087** |
| PrecWD with 20 Ordinal Encoding | 195.750*** | 1.319*** | 2.492* | 0.037* | 193.992*** | 1.294*** |
| PrecWD without Attention | 175.104* | 1.053* | 2.468* | 0.037* | 174.582* | 1.046* |
| **All Features** | **165.442** | **0.992** | **2.411** | **0.031** | **165.296** | **0.991** |
| Without Webpage | 189.535*** | 1.153** | 2.529** | 0.035* | 189.535*** | 1.153** |
| Without Unstructured | 182.847*** | 1.086** | 2.492* | 0.037* | 182.539*** | 1.084** |
| Without Acoustic | 177.610** | 1.030 | 2.484* | 0.036* | 177.158** | 1.033* |
| Without Description | 174.599* | 1.018 | 2.455* | 0.034 | 175.551** | 1.027 |
| Without Transcript | 170.482* | 1.004 | 2.489* | 0.036* | 169.540* | 0.998 |
| Without Channel | 170.391* | 1.006 | 2.493* | 0.037* | 169.574* | 1.000 |

*: $p < 0.05$; **: $p < 0.01$; ***: $p < 0.001$

## Table 7. Evaluation of WGAN-GP

| Metric | Model | Comp 1 | Comp 2 | Comp 3 | Comp 4 | Comp 5 | Comp 6 | Comp 7 | Comp 8 | Comp 9 | Comp 10 |
|---|---|---|---|---|---|---|---|---|---|---|---|
| Mean | Real | -2.091 | 0.434 | 0.007 | -0.104 | -0.157 | -0.257 | 0.031 | 0.032 | -0.194 | -0.039 |
| | WGAN-GP | -2.073 | 0.592 | 0.177 | -0.295 | -0.317 | -0.172 | 0.018 | 0.051 | -0.057 | -0.031 |
| | VAE | 2.068*** | -0.370* | -0.410 | 0.296 | 0.190 | 0.243* | -0.013 | 0.003 | 0.153* | 0.045 |
| | Bayesian | 2.096*** | -0.656*** | 0.226 | 0.103 | 0.284* | 0.185** | -0.036 | -0.086 | 0.097 | 0.025 |
| Variance | Real | 0.528 | 0.751 | 0.837 | 0.551 | 0.474 | 0.423 | 0.403 | 0.476 | 0.347 | 0.323 |
| | WGAN-GP | 0.706 | 0.795 | 0.707 | 0.531 | 0.457 | 0.509 | 0.503 | 0.447 | 0.473 | 0.335 |
| | VAE | 1.825*** | 2.509** | 1.619 | 1.845 | 1.476* | 1.526* | 1.370 | 1.293 | 1.161 | 0.876** |
| | Bayesian | 2.067*** | 2.321*** | 2.328 | 1.644 | 1.724** | 1.479* | 1.441 | 1.312 | 1.367 | 1.436 |

Note: The significance tests are comparing the generated samples with the real samples. Not significant means no difference from real samples.
*: $p < 0.05$; **: $p < 0.01$; ***: $p < 0.001$

## Table 8. Precise Interpretation (Total Effect) v.s. Existing Approach (Main Effect) (Normalized)

| Feature | Total Effect | Main Effect | Linear Regression (no interaction) | Linear Regression (interaction) | PrecWD's Main Effect (remove higher-order) |
|---|---|---|---|---|---|
| Average Bitrate | -0.135 | -0.151 | 3.252 | 4.296 | -0.025 |
| Description Bigrams | -0.020 | -0.853 | -0.131 | -0.782 | -0.183 |
| Use Cipher | 0.000 | 0.053 | 0.000 | 0.750 | 0.028 |
| Audio Sample Rate | 0.000 | -0.459 | 0.000 | 0.757 | -0.014 |
| Appearance of Numbers in Transcript | 0.000 | 1.952 | 0.000 | -0.042 | 0.584 |
| Is Organization | 0.000 | 1.852 | 0.000 | **-0.055** | 1.085 |
| Appearance of Numbers in Description | 0.025 | 0.200 | 0.683 | 0.006 | 0.041 |
| Transcript Length | 0.084 | 0.009 | -1.906 | 0.076 | 0.039 |
| Description Length | 0.086 | 5.882 | 1.899 | 0.006 | 2.577 |
| Transcript Informativeness | 0.105 | **-0.200** | 0.029 | **-0.003** | **-0.018** |
| Transcript Bigrams | 0.109 | -0.816 | -0.029 | 0.036 | -0.597 |
| Transcript Neutral Sentiment | 0.140 | 0.759 | -0.771 | 0.061 | 0.859 |
| Description Positive Sentiment | 0.259 | 4.209 | 0.448 | -0.018 | 1.581 |
| Transcript Positive Sentiment | 0.301 | -0.882 | 0.525 | -0.001 | -0.550 |
| Allow Ratings | 0.358 | 0.659 | **-0.257** | 0.005 | 0.295 |
| Video Duration | 0.384 | 0.100 | -0.732 | 0.061 | 0.158 |
| Transcript Readability | 0.390 | **-2.959** | 0.414 | **-0.007** | 0.049 |
| Description Informativeness | 0.396 | 1.043 | 0.044 | 0.030 | 0.527 |
| Description Neutral Sentiment | 0.400 | 5.288 | 1.660 | -0.004 | 0.772 |
| Is Verified | 0.443 | **-0.221** | **-1.661** | **-0.023** | **-0.200** |
| Transcript Compound Sentiment | 0.503 | **-0.186** | **-0.352** | **-0.002** | **-0.583** |
| Description Med Knowledge | 0.530 | 0.496 | 0.001 | 0.007 | 1.284 |
| Description Vocabulary Richness | 0.533 | 2.594 | -1.913 | -0.011 | 1.588 |
| No. Audio Tracks | 0.539 | **-0.053** | 0.000 | 0.008 | **-0.089** |
| Is Live Content | 0.539 | 0.858 | -0.396 | -0.001 | -0.007 |
| Description Negative Sentiment | 0.595 | 2.594 | -0.968 | 0.001 | 1.583 |
| Description Compound Sentiment | 0.690 | 0.099 | **-0.858** | 0.000 | 0.005 |
| Transcript Negative Sentiment | 0.784 | 0.120 | -0.291 | 0.000 | 0.408 |
| Transcript Vocabulary Richness | 0.834 | 1.160 | -0.006 | 0.001 | 1.195 |
| Transcript Med Knowledge | 0.940 | **-1.588** | 0.001 | **-0.001** | **-0.553** |
| Description Readability | 0.989 | **-3.959** | 3.443 | **-0.001** | **-1.747** |

Note: The bolded ones are non-exclusive examples whose signs are different from common perception. The feature importance rankings of the total effect and other approaches are also significantly different.

### Table 9. Prediction Comparison of PrecWD with Baseline Models (Case Study 2)

| Method | Outcome variable: ADV | | Outcome Variable: log(total view) (add published_days as a feature) | | Outcome Variable: ADV (add published_days as a feature) | |
|---|---|---|---|---|---|---|
| | MSE | MSLE | MSE | MSLE | MSE | MSLE |
| **PrecWD (Ours)** | **140.202** | **0.728** | **2.156** | **0.022** | **139.583** | **0.713** |
| Linear regression | 881.027*** | 3.184*** | 2.453*** | 0.056*** | 880.529*** | 3.180*** |
| KNN-1 | 227.479*** | 2.421*** | 3.445*** | 0.056*** | 222.100*** | 2.392*** |
| KNN-3 | 163.061** | 2.327*** | 2.354*** | 0.039*** | 164.580*** | 2.281*** |
| KNN-5 | 180.479*** | 2.264*** | 2.317*** | 0.038*** | 180.583*** | 2.277*** |
| DT-MSE | 284.387*** | 3.362*** | 4.082*** | 0.061*** | 280.491*** | 3.365*** |
| DT-MAE | 288.223*** | 3.193*** | 3.898*** | 0.059*** | 285.224*** | 3.204*** |
| DT-Fredmanmse | 299.519*** | 3.435*** | 3.771*** | 0.060*** | 295.186*** | 3.402*** |
| SVR-Linear | 185.644*** | 4.924*** | 3.503*** | 0.052*** | 183.584*** | 4.885*** |
| SVR-RBF | 185.989*** | 4.901*** | 3.539*** | 0.052*** | 186.351*** | 4.923*** |
| SVR-Poly | 192.951*** | 4.910*** | 3.898*** | 0.057*** | 190.740*** | 4.905*** |
| SVR-Sigmoid | 185.646*** | 4.924*** | 3.494*** | 0.052*** | 185.438*** | 4.913*** |
| Gaussian Process-1 | 1290.452*** | 8.487*** | 3.057*** | 5.974*** | 1278.945*** | 8.469*** |
| Gaussian Process-3 | 1287.329*** | 8.446*** | 4.327*** | 6.038*** | 1274.183*** | 8.459*** |
| Gaussian Process-5 | 1283.331*** | 8.439*** | 4.337*** | 6.046*** | 1272.429*** | 8.420*** |
| MLP-1 | 172.460*** | 1.005*** | 2.245** | 0.029* | 172.584*** | 1.005*** |
| MLP-2 | 169.147*** | 1.063*** | 2.334*** | 0.033** | 170.291*** | 1.074*** |
| MLP-3 | 162.249** | 0.950** | 2.349*** | 0.030* | 162.692** | 0.982** |
| MLP-4 | 181.192*** | 0.932** | 2.318*** | 0.031* | 181.583*** | 0.941** |
| CNN-1 | 245.382*** | 1.377*** | 2.235** | 0.028** | 244.592*** | 1.358*** |
| CNN-2 | 158.040** | 1.090*** | 2.244** | 0.029** | 158.206** | 1.024*** |
| CNN-3 | 155.651* | 1.023*** | 2.249** | 0.026* | 156.583** | 1.050*** |
| CNN-4 | 169.584** | 1.065*** | 2.251** | 0.029* | 167.344** | 1.068*** |
| LSTM-1 | 341.301*** | 1.718*** | 2.385*** | 0.033** | 342.590*** | 1.738*** |
| LSTM-2 | 182.828*** | 1.099*** | 2.347*** | 0.031** | 185.493*** | 1.184*** |
| BLSTM-1 | 367.261*** | 1.661*** | 2.356*** | 0.028* | 370.194*** | 1.672*** |
| BLSTM-2 | 175.995*** | 0.999*** | 2.290*** | 0.026* | 175.143*** | 0.999** |
| W&D | 180.869*** | 1.067** | 2.210* | 0.025* | 179.528*** | 1.056*** |
| W&D-CNN | 186.773*** | 1.866*** | 2.223* | 0.027* | 184.193*** | 1.859*** |
| W&D-LSTM | 183.719*** | 1.598*** | 2.229* | 0.027* | 184.511*** | 1.740*** |
| W&D-BLSTM | 206.321*** | 2.454*** | 2.231* | 0.027* | 203.776*** | 2.406*** |
| Piecewise W&D-10 | 227.633*** | 3.116*** | 2.359*** | 0.035** | 219.261*** | 3.055*** |
| Piecewise W&D-20 | 206.792*** | 3.016*** | 2.362*** | 0.039*** | 205.147*** | 2.987*** |

*: $p < 0.05$; **: $p < 0.01$; ***: $p < 0.001$

**Table 10. Ablation Studies (Case Study 2)**

| Method | Outcome variable: ADV | | Outcome Variable: log(total view) (add published_days as a feature) | | Outcome Variable: ADV (add published_days as a feature) | |
|---|---|---|---|---|---|---|
| | MSE | MSLE | MSE | MSLE | MSE | MSLE |
| **PrecWD** | **140.202** | **0.728** | **2.156** | **0.022** | **139.583** | **0.713** |
| PrecWD without Unstructured Component | 151.259* | 0.887* | 2.257** | 0.028* | 150.582* | 0.884* |
| PrecWD without Piecewise Linear Component | 175.136** | 0.915* | 2.384*** | 0.033** | 175.249** | 0.920* |
| PrecWD without Second-Order Component | 155.984* | 0.890* | 2.300*** | 0.031** | 156.538* | 0.894* |
| PrecWD without High-order Component | 159.354* | 0.848 | 2.296** | 0.029* | 159.000* | 0.841* |
| PrecWD with Simple Linear Encoding | 153.454* | 0.872** | 2.204* | 0.025* | 153.119* | 0.870* |
| PrecWD with 10 Ordinal One-hot Encoding | 167.449** | 0.938** | 2.297** | 0.028* | 165.838** | 0.923*** |
| PrecWD with 20 Ordinal One-hot Encoding | 168.621** | 0.968*** | 2.284** | 0.028* | 167.206** | 0.971*** |
| PrecWD with 10 Ordinal Encoding | 195.510*** | 0.862*** | 2.319*** | 0.031** | 195.924*** | 0.870*** |
| PrecWD with 20 Ordinal Encoding | 195.510*** | 0.862*** | 2.303*** | 0.029** | 195.924*** | 0.870*** |
| PrecWD without Attention | 153.454* | 0.872** | 2.227* | 0.024* | 153.829* | 0.869*** |
| **Data Sources** | **MSE** | **MSLE** | **MSE** | **MSLE** | **MSE** | **MSLE** |
| **All (Ours)** | **140.202** | **0.728** | **2.156** | **0.022** | **139.583** | **0.713** |
| Without Webpage | 194.553*** | 0.908** | 2.219* | 0.030* | 194.553*** | 0.908** |
| Without Unstructured | 164.936*** | 0.779 | 2.202* | 0.025* | 164.821*** | 0.776 |
| Without Acoustic | 155.798** | 0.817* | 2.197* | 0.026* | 154.588** | 0.801* |
| Without Description | 156.647** | 0.832* | 2.209* | 0.026* | 157.024** | 0.846* |
| Without Transcript | 149.663* | 0.770 | 2.214* | 0.028* | 148.763* | 0.765 |
| Without Channel | 147.250* | 0.737 | 2.224* | 0.028* | 146.217* | 0.720 |

*: $p < 0.05$; **: $p < 0.01$; ***: $p < 0.001$

**Table 11. Evaluation of WGAN-GP (Case Study 2)**

| Metric | Model | Comp 1 | Comp 2 | Comp 3 | Comp 4 | Comp 5 | Comp 6 | Comp 7 | Comp 8 | Comp 9 | Comp 10 |
|---|---|---|---|---|---|---|---|---|---|---|---|
| Mean | Real | -2.077 | -0.489 | 0.609 | -0.220 | 0.046 | 0.030 | -0.026 | 0.020 | 0.008 | 0.002 |
| | WGAN-GP | -2.111 | -0.520 | 0.592 | -0.230 | 0.042 | 0.028 | -0.017 | 0.008 | 0.018 | 0.002 |
| | VAE | -2.087 | -0.559* | 0.597 | -0.215 | 0.016 | 0.002* | -0.017 | 0.010 | 0.007 | -0.001 |
| | Bayesian | 6.274*** | 1.568*** | -1.798*** | 0.664*** | -0.104 | -0.060 | 0.060 | -0.038 | -0.032 | -0.003 |
| Variance | Real | 0.328 | 0.173 | 0.112 | 0.080 | 0.080 | 0.044 | 0.026 | 0.017 | 0.010 | 0.022 |
| | WGAN-GP | 0.304 | 0.162 | 0.131 | 0.083 | 0.074 | 0.048 | 0.025 | 0.013 | 0.010 | 0.025 |
| | VAE | 0.118 | 0.111* | 0.051 | 0.043 | 0.034 | 0.017* | 0.004 | 0.006 | 0.004 | 0.015 |
| | Bayesian | 35.802*** | 29.417*** | 15.080*** | 9.004*** | 5.245 | 4.356 | 3.740 | 3.285 | 3.073 | 2.847 |

Note: The significance tests are comparing the generated samples with the real samples. Not significant means no difference from real samples.
*: $p < 0.05$; **: $p < 0.01$; ***: $p < 0.001$

**Table 12. User Study Groups**

| Group | Model | Rationale |
|---|---|---|
| A | PrecWD | Our model |
| B | W&D | Best-performing interpretable baseline |
| C | Piecewise W&D | State-of-the-art model-based interpretable model |
| D | SHAP (using our prediction model) | State-of-the-art *post-hoc* explanation model |
| E | VAE-based model | Best-performing generative baseline |

**Table 13: Interpretability Comparison of PrecWD and Interpretable Methods**

| Group | Mean of Interpretability: Trust | Mean of Interpretability: Usefulness |
|---|---|---|
| PrecWD | 2.183 | 2.101 |
| W&D | 1.820 * | 1.810 * |
| Piecewise W&D | 0.676 *** | 0.548 *** |
| SHAP | 1.429 *** | 1.304 *** |
| VAE | 1.202 *** | 1.160 *** |

**Table 14. Number of Participants in Original and Final Model**

| | Original: PrecWD | Original: Baseline |
|---|---|---|
| Switch to PrecWD | 21 | 84 |
| Switch to Baseline | 7 | 28 |

**Table 15. Comparison of 1st and 2nd Time Interpretability**

| | Interpretability Measurement | | | |
|---|---|---|---|---|
| | Mean of 1st Time Trust | Mean of 2nd Time Trust | Mean of 1st Time Usefulness | Mean of 2nd Time Usefulness |
| PrecWD → PrecWD | 2.418 | 2.635* | 2.286 | 2.500 |
| Baseline → PrecWD | 1.054 | 2.217*** | 0.996 | 2.219*** |

**Table 16: User Study 2**

| Group | Mean of Interpretability: Trust | Mean of Interpretability: Usefulness |
|---|---|---|
| PrecWD | 2.080 | 2.007 |
| SHAP | 1.470 ** | 1.300 ** |

## Figures

**Figure 1. PrecWD Architecture**

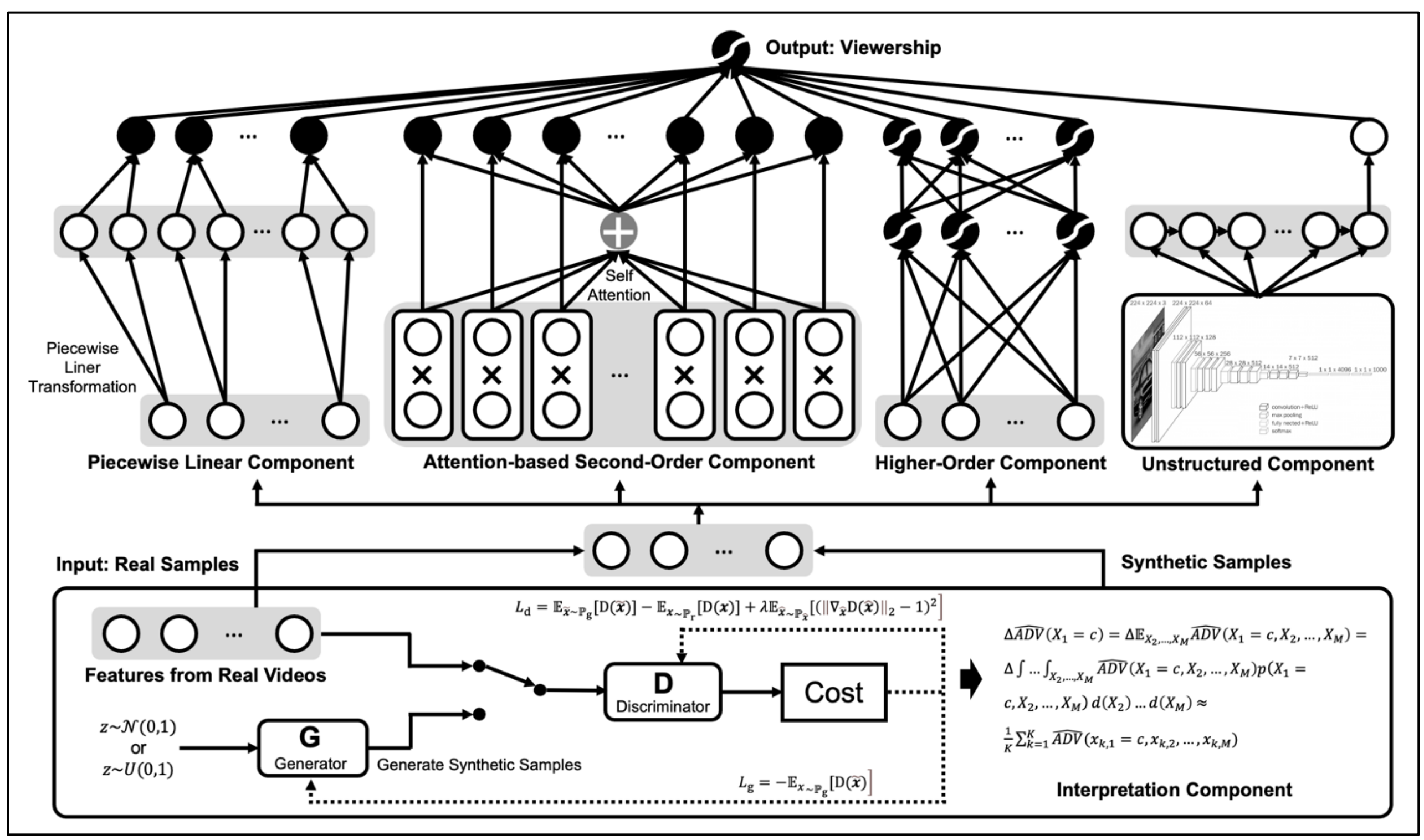

**Figure 2. WGAN-GP Convergence**

**2.a. Discriminator/Generator Loss** **2.b. Train/Valid Loss**

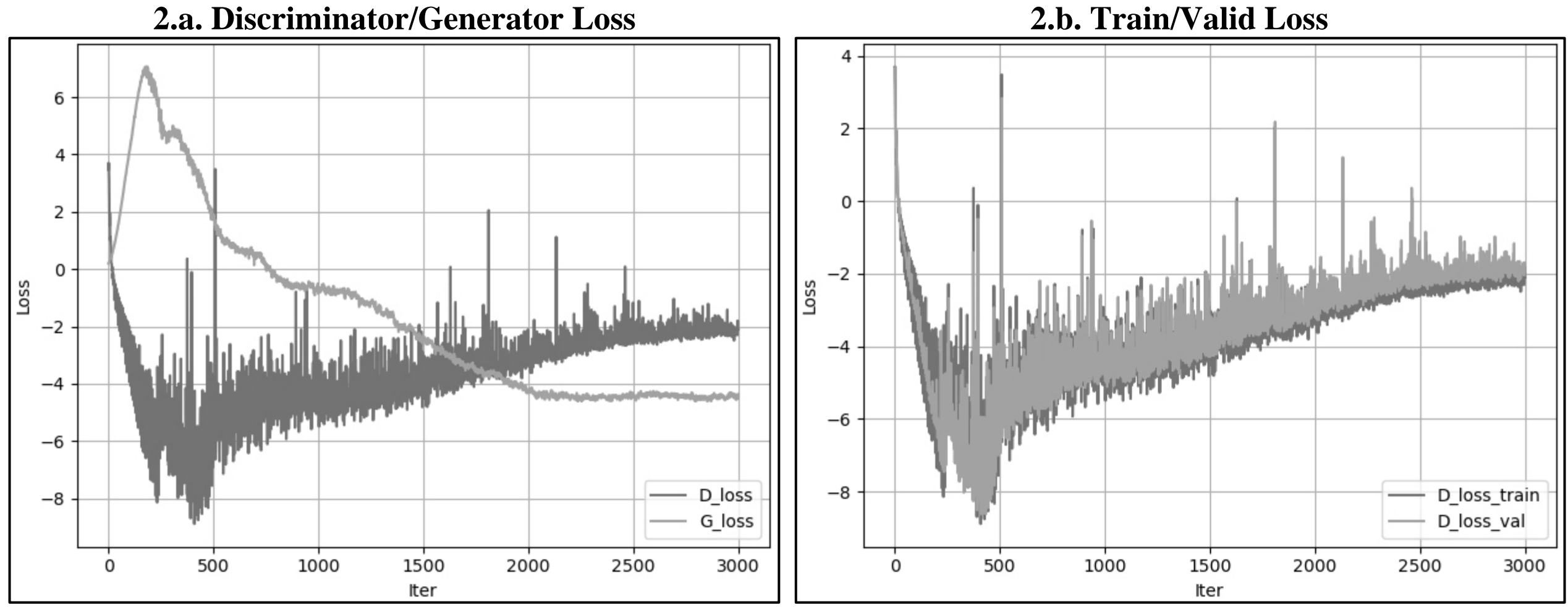

Note: The Wasserstein loss requires using y = 1 and -1, rather than 1 and 0. Therefore, WGAN-GP removes the sigmoid activation from the final layer of the discriminator, so that predictions are no longer constrained to [0,1], but instead can be $(-\infty, \infty)$ (https://www.oreilly.com/library/view/generative-deep-learning/9781492041931/ch04.html).

**Figure 3. Generated Samples Evaluation**

**3.a. Sample Distribution** **3.b. Description Readability**

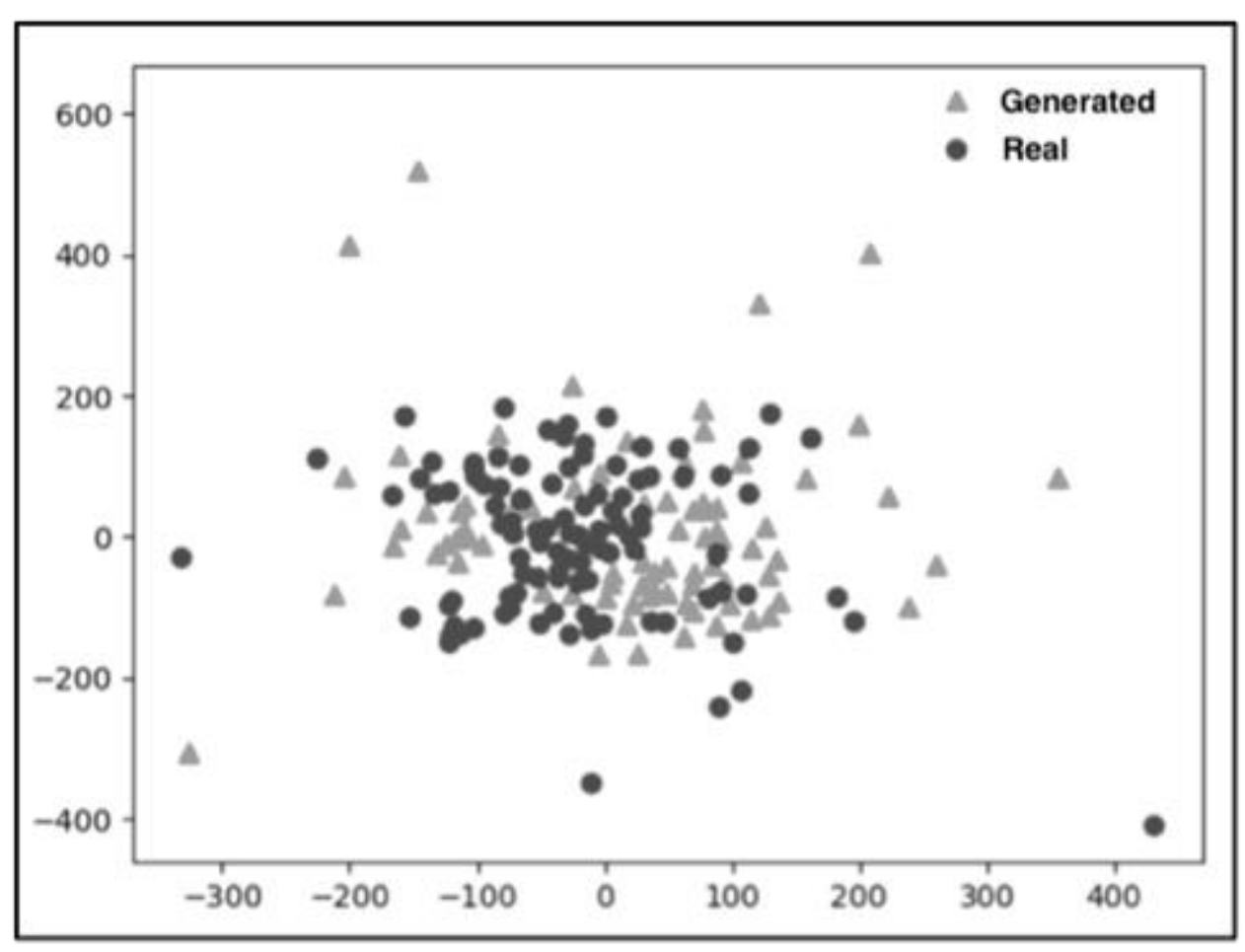

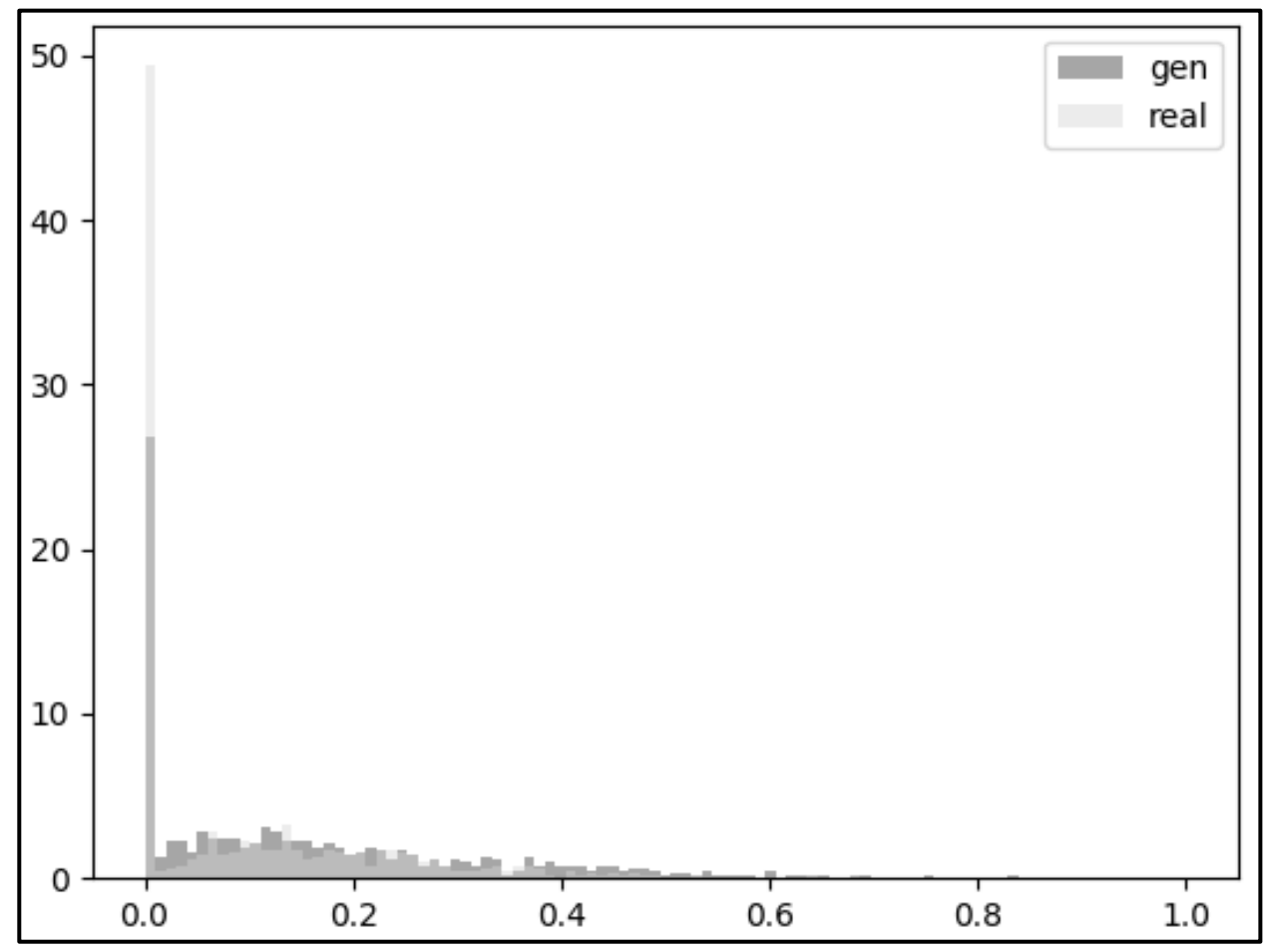

**3.c. Transcript Medical Knowledge**

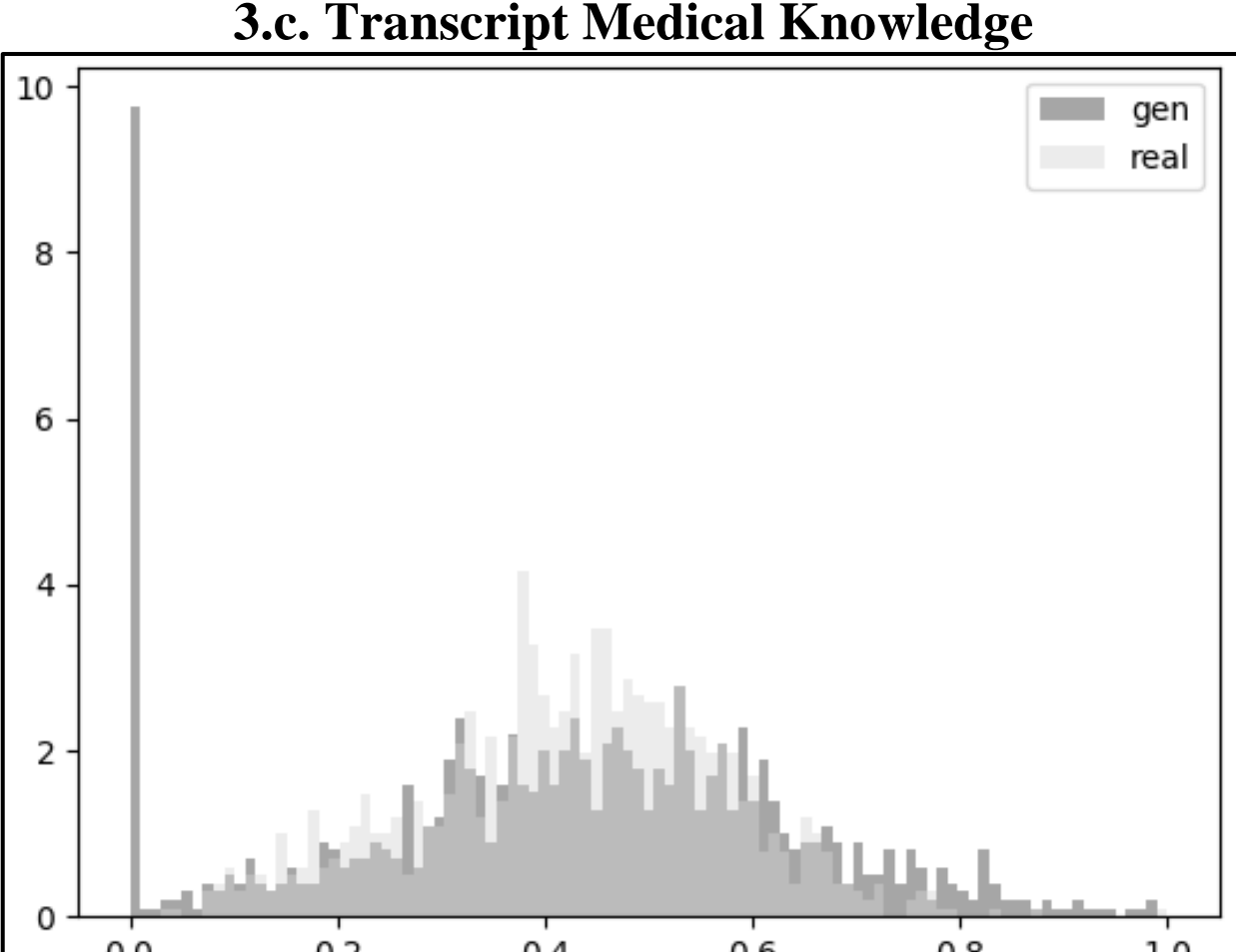

**Figure 4. Feature-based Interpretation (Normalized)**

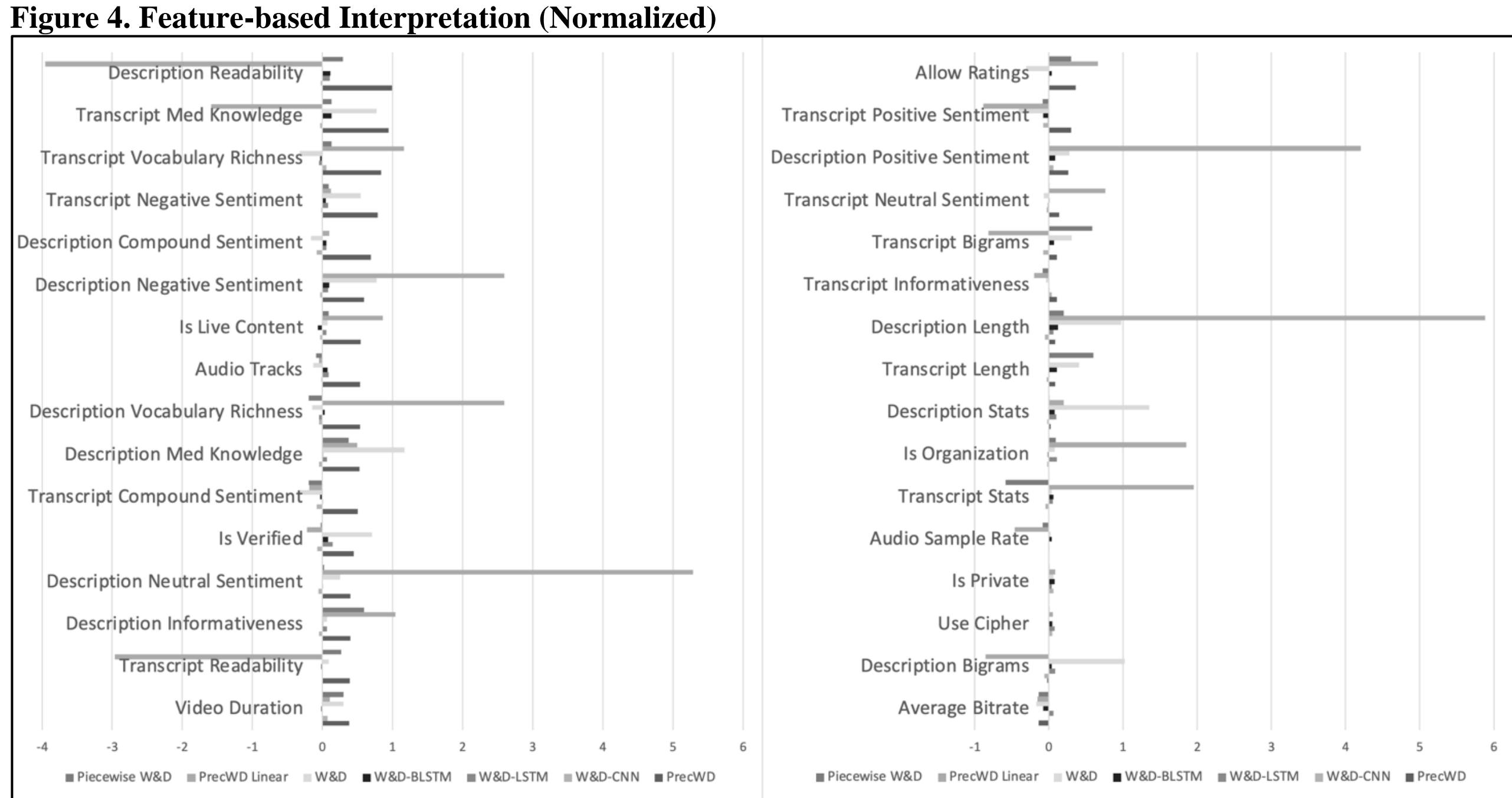

Note: To compare the features in the same scale, we normalized the effect values. The x-axis is the weight of a variable. The higher the absolute value of the weight is, the more important the variable is.

**Figure 5. Examples of the Dynamic Total Effect**

**5.a. Description Vocabulary Richness**

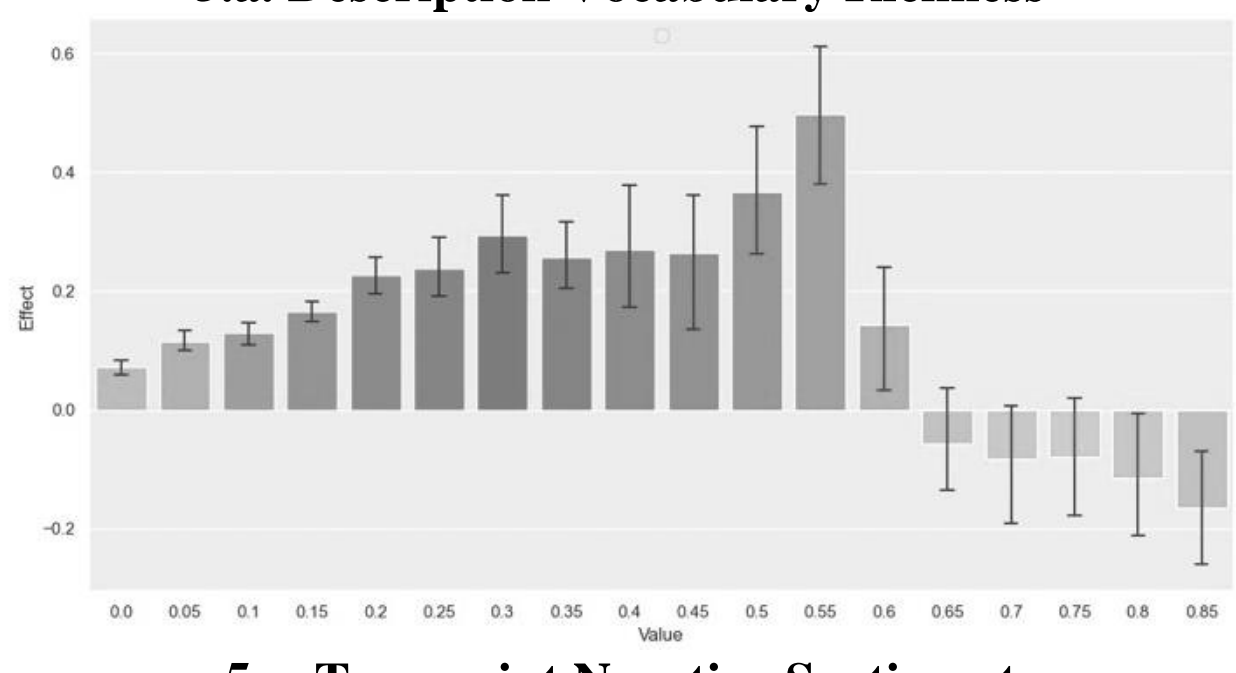

**5.b. Description Readability**

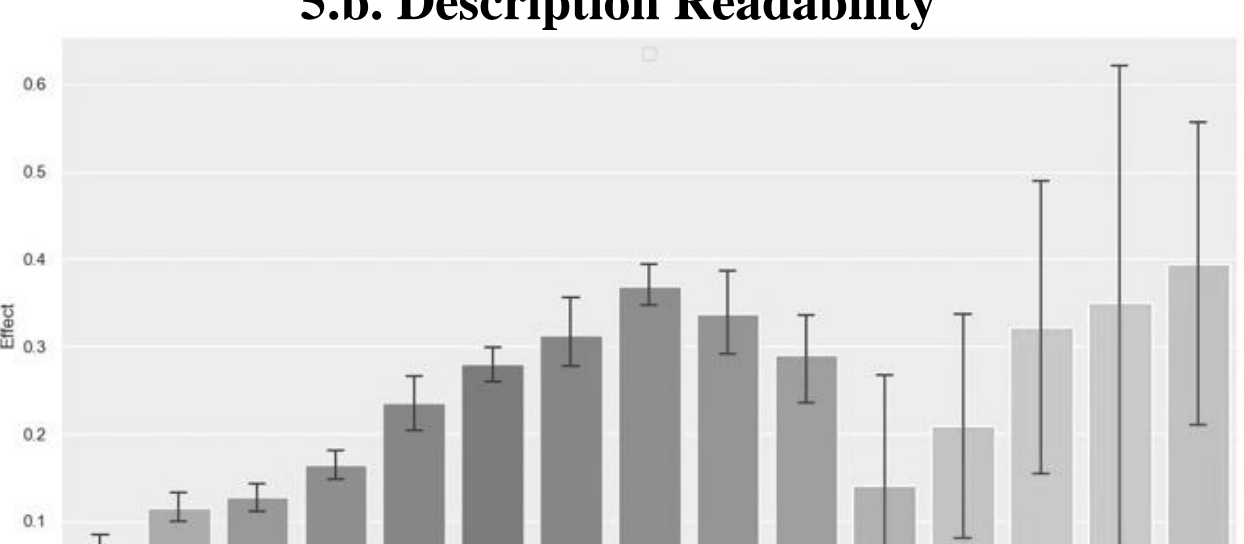

**5.c. Transcript Negative Sentiment**

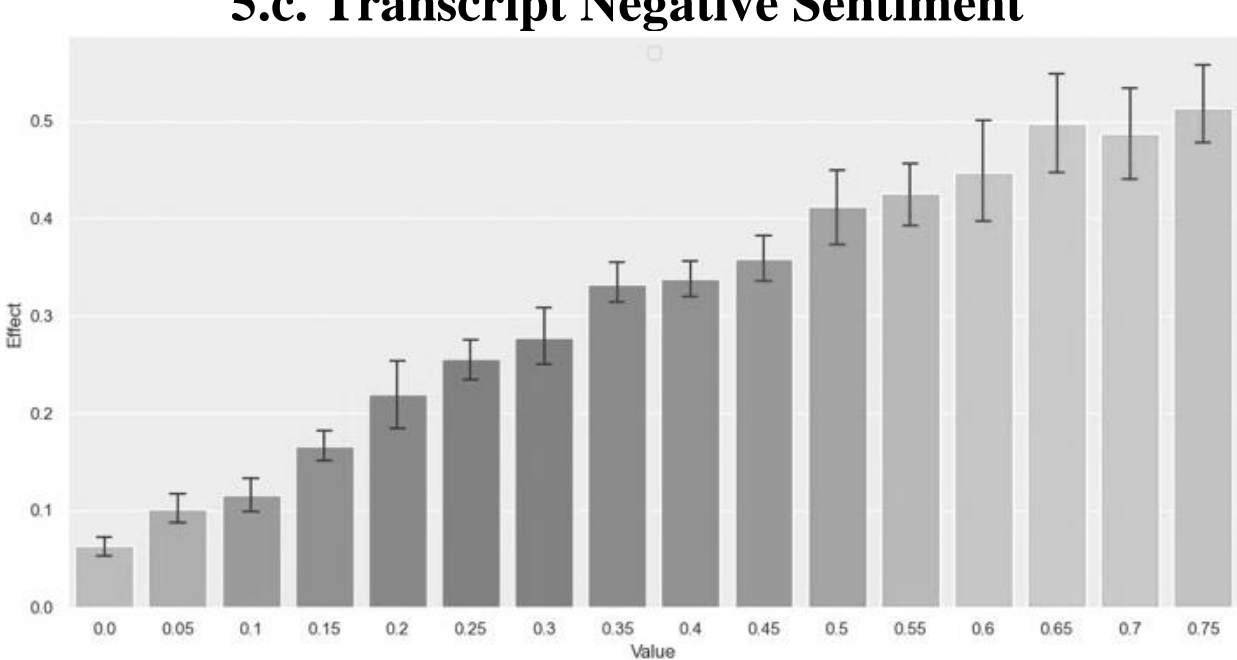

**Figure 6. Our Model in User Study 1**

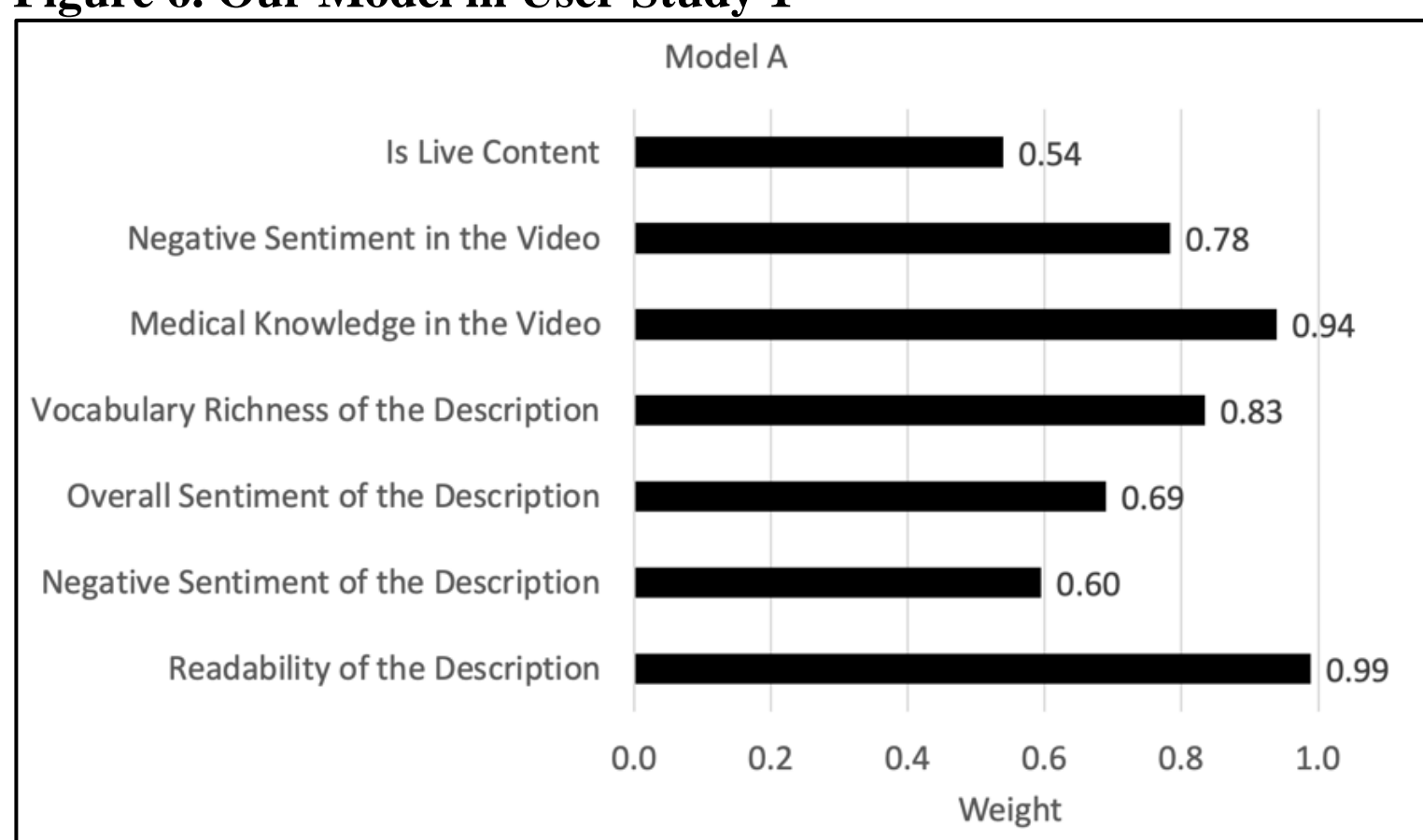

**Figure 7. Models in User Study 2 (Left: PrecWD, Right: SHAP)**

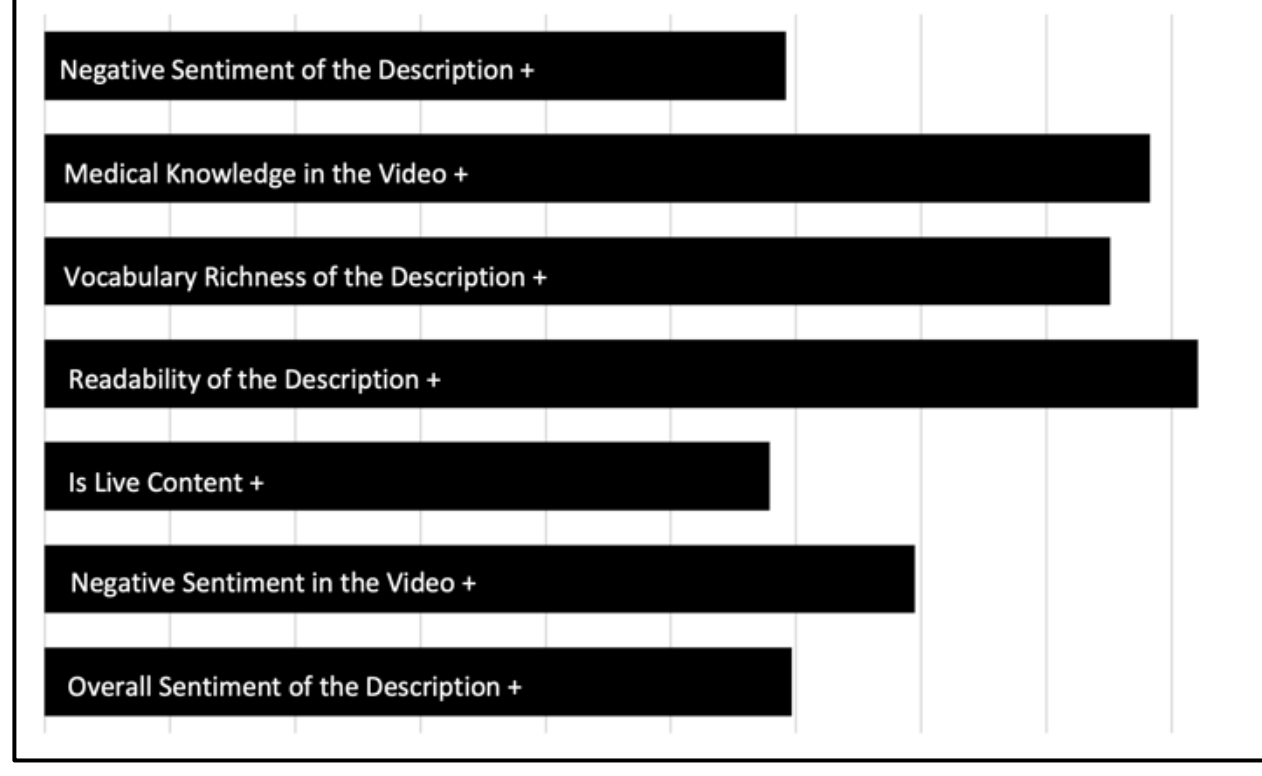

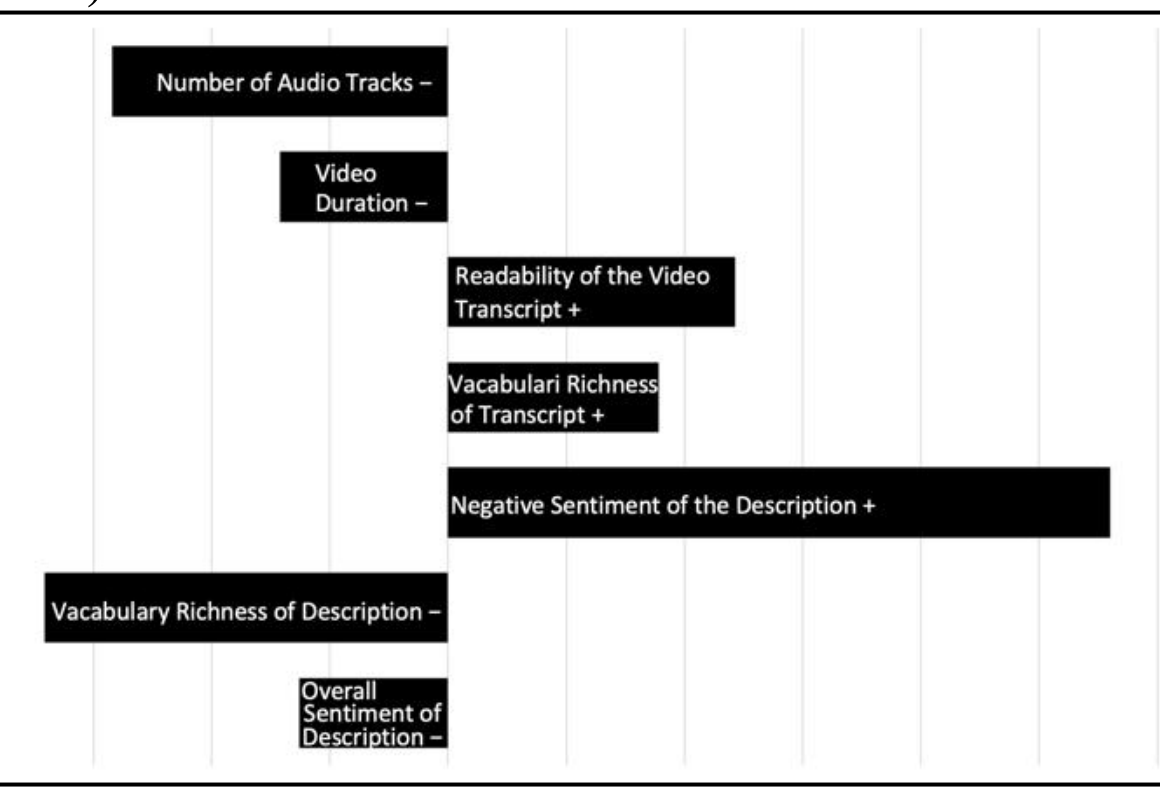